\documentclass[10pt,conference, letterpaper]{ IEEEtran}
\IEEEoverridecommandlockouts
\makeatletter
\newcommand{\linebreakand}{%
  \end{@IEEEauthorhalign}
  \hfill\mbox{}\par
  \mbox{}\hfill\begin{@IEEEauthorhalign}
}
\makeatother

\usepackage{pifont}
\usepackage{algorithm}
\usepackage{algorithmic}

\usepackage{hyperref}

\usepackage{tabularx}
\usepackage{amsmath}
\usepackage{amssymb}
\usepackage{graphicx}
\usepackage{booktabs}  
\usepackage{threeparttable} \usepackage{multirow} 
\usepackage{color}

\usepackage[caption=false,font=footnotesize]{subfig}

\usepackage{cite}
\usepackage{amsmath,amssymb,amsfonts}
\usepackage{algorithmic}
\usepackage{graphicx}
\usepackage{textcomp}
\usepackage{xcolor}

\newtheorem{theorem}{Theorem}

\newtheorem{lemma}{Lemma}

\newtheorem{definition}{Definition}
\newtheorem{assumption}{Assumption}

\def\BibTeX{{\rm B\kern-.05em{\sc i\kern-.025em b}\kern-.08em
    T\kern-.1667em\lower.7ex\hbox{E}\kern-.125emX}}
\begin{document}

\title{First-order Constrained Trilevel Optimization
Over Distributed Networks for Robust Coreset
Selection
}

\author{\IEEEauthorblockN{Yang Jiao}
\IEEEauthorblockA{\textit{Southeast University}\\
yang.jiao@seu.edu.cn} \\
\and
\IEEEauthorblockN{Kaixuan Jiao}
\IEEEauthorblockA{\textit{Anhui University}\\
kaixuanjiao@stu.ahu.edu.cn} \\
\and
\IEEEauthorblockN{Kai Yang}
\IEEEauthorblockA{\textit{Tongji University}\\
kaiyang@tongji.edu.cn}\\ 
\linebreakand 
\IEEEauthorblockN{Nadjib Aitsaadi}
\IEEEauthorblockA{\textit{UVSQ Paris-Saclay $\&$ DAVIDLab}\\
nadjib.aitsaadi@uvsq.fr} \\
\and
\IEEEauthorblockN{Ilhem Fajjari}
\IEEEauthorblockA{\textit{Orange Innovation}\\
ilhem.fajjari@orange.com} \\
\and
\IEEEauthorblockN{Renwei (Richard) Li}
\IEEEauthorblockA{\textit{Southeast University}\\
richard.li@seu.edu.cn} \\
% \and

% \author{\IEEEauthorblockN{1\textsuperscript{st} Yang Jiao}
% \IEEEauthorblockA{\textit{Southeast University}} \\
% \and
% \IEEEauthorblockN{2\textsuperscript{nd} Kaixuan Jiao}
% \IEEEauthorblockA{\textit{Anhui University}} \\
% \and
% \and
% \and
% \and
}

\maketitle

\begin{abstract}
With the rapid advancement of the Internet of Things (IoT), massive amounts of data are generated across distributed edge networks. Training models on full data incurs significant computational overhead and storage bottlenecks, rendering coreset selection a critical paradigm. Furthermore, given the privacy-sensitive nature of local data and the escalating demand for model robustness in real-world deployments, developing an effective distributed optimization framework for robust coreset selection is vital, yet remains largely unexplored. To this end, this work first characterizes the hierarchical dependencies among coreset selection, robust optimization, and distributed learning, and formulates the distributed robust coreset selection as a trilevel optimization problem with level-wise constraints. Furthermore, to effectively solve the trilevel problem in a distributed manner, the \underline{F}ederated \underline{F}irst-order \underline{C}onstrained \underline{T}rilevel \underline{O}ptimization (F$^2$CTO) is proposed, which synergistically integrates a hierarchical composite value-function reformulation and a distributed alternating projected gradient algorithm. To the best of our knowledge, F$^2$CTO is the first method developed for distributed robust coreset selection, as well as the first distributed optimization approach for trilevel optimization problems with level-wise constraints. Additionally, we prove that the proposed method achieves a non-asymptotic convergence rate of $\mathcal{O}(\epsilon^{-3/2})$ for finding an $\epsilon$-stationary point. Extensive empirical evaluations on reliable continual learning demonstrate the effectiveness and efficiency of the proposed F$^2$CTO.

\end{abstract}


\section{Introduction}

Driven by the rapid development of the Internet of Things (IoT), massive volumes of data are continuously generated and dispersed across nodes over distributed networks \cite{wang2018edge}. Training machine learning models directly on the entirety of this data incurs prohibitive computational overhead and severe storage bottlenecks \cite{xia2022moderate}. To alleviate these training and memory constraints, coreset selection has garnered significant attention \cite{moser2025coreset}. This technique involves extracting a representative and information-dense subset (i.e., the coreset) from a large-scale dataset, such that a model trained on this subset can closely approximate the performance of one trained on the entire dataset. Coreset selection has been widely used across various fields because of its remarkable ability to alleviate computational and storage bottlenecks while preserving the informational integrity of large-scale datasets. For instance, to reduce the computational costs associated with training large language models (LLMs) on massive corpora, coreset selection has been employed to extract representative subsets of instruction data, thereby significantly improving training efficiency \cite{shen2026brief}. Additionally, to enable models deployed in Internet of Things (IoT) environments to acquire new knowledge without suffering from catastrophic forgetting, coreset selection has been leveraged to facilitate continual learning \cite{chen2025online}. Moreover, in the field of network security, coreset selection has been utilized to reduce memory and storage requirements, ultimately improving the efficiency of Bayesian learning for network intrusion detection \cite{zennaro2019analyzing}.

In addition to computational and memory constraints, the highly sensitive nature of local data in distributed networks poses a significant challenge. Centralizing this data not only risks severe privacy breaches but also incurs massive communication overhead \cite{jiao2024provably}, thereby necessitating the development of distributed optimization methods for coreset selection. Moreover, as the demand for reliable machine learning models grows, extracting critical data from large-scale datasets to enhance model robustness, thereby achieving efficient robust optimization, has attracted considerable research attention \cite{jiao2025aspire}. However, existing coreset selection methods either overlook the robustness of coresets or rely on centralized setups, which limit their applicability to distributed robust coreset selection. In light of these limitations, the first key question this paper aims to address is: \textit{Can we construct an effective distributed robust coreset selection framework?} Such a framework is expected to enable individual nodes in a network to identify representative data without sharing raw data, while more effectively leveraging the global information through coordination by a master node. The motivating applications of studying distributed robust coreset selection are provided in Sec. \ref{sec:motivating}.

Coreset selection is typically formulated as a bilevel optimization problem \cite{hao2023bilevel}, where the lower-level optimization trains model parameters based on the selected coreset, and the upper-level optimization evaluates these parameters to further refine the coreset selection. Building upon this paradigm and investigating the inherent coupling among coreset selection, robust optimization, and distributed learning, distributed robust coreset selection can be formulated as a constrained trilevel optimization problem. Compared to trilevel optimization problems without additional constraints, which are the primary focus of existing literature \cite{jiao2024provably,giovannelli2025stochastic}, this constrained variant is harder to solve, leaving the corresponding distributed algorithms largely unexplored. Motivated by this gap, this work seeks to answer a second key question: \textit{Can we design an efficient distributed first-order algorithm with non-asymptotic convergence guarantees to solve the constrained trilevel optimization problem?}

To this end, we propose a \underline{F}ederated \underline{F}irst-order \underline{C}onstrained \underline{T}rilevel \underline{O}ptimization framework, termed F$^2$CTO, for distributed robust coreset selection. Within this framework, we first investigate the underlying coupling inherent in robust coreset selection and formulate it as a trilevel optimization problem with level-wise constraints. Subsequently, a hierarchical composite value-function method is proposed to transform the original optimization problem into a formulation amenable to distributed optimization, comprising one inner-level and two outer-level value-functions. Building upon this, we introduce a distributed first-order alternating projected gradient algorithm to efficiently address the resulting problem. To the best of our knowledge, this is the first work to address the distributed robust coreset selection problem, and the first distributed algorithm designed to solve constrained trilevel optimization problems. Theoretically, we analyze the non-asymptotic convergence rate of the proposed method to achieve an $\epsilon$-stationary point, explicitly characterizing both its iteration and communication complexities. The contributions of this work are summarized as follows:

\noindent \textbf{1.} Unlike existing coreset selection works that either focus exclusively on centralized settings or overlook the necessity of robustness, this work takes an initial step toward integrating coreset selection, robust optimization, and distributed learning into a unified trilevel optimization framework to address the distributed robust coreset selection problem.

\noindent \textbf{2.} Compared to existing trilevel optimization methods, to the best of our knowledge, the proposed F$^2$CTO is the first to address trilevel optimization with level-wise constraints in a distributed manner. Theoretically, we demonstrate that the non-asymptotic convergence rate for the proposed F$^2$CTO to achieve an $\epsilon$-stationary point is upper bounded by $\mathcal{O}(\epsilon^{-3/2})$.

\noindent \textbf{3.}  Extensive experiments on reliable continual learning demonstrate the superior effectiveness and efficiency of F$^2$CTO in addressing distributed robust coreset selection.

\section{Related Work}

\subsection{Coreset Selection}

Coreset selection aims to extract a representative subset $\mathcal{D}^*$ from a large-scale dataset $\mathcal{D}$ ($\mathcal{D}^* \!\subset \! \mathcal{D}$) to improve training efficiency and reduce memory overhead. As discussed in \cite{moser2025coreset}, existing coreset selection methods can be broadly categorized into training-free \cite{xiao2024feature}, scoring \cite{xia2022moderate}, decision boundary \cite{yang2024mind}, submodularity \cite{karanam2022orient}, gradient matching \cite{killamsetty2021grad}, and bilevel optimization \cite{shen2026brief} approaches. As an alternative to coreset selection, data distillation aims to synthesize a compact dataset but is highly sensitive to model architecture \cite{zhou2022probabilistic}, making a direct performance comparison with coreset selection inappropriate \cite{xia2022moderate}.
In addition, recent studies have demonstrated the promise of robust coreset selection in enhancing the robustness of machine learning models \cite{dolatabadi2023adversarial,shinde2025data}. Furthermore, in real-world scenarios, data is frequently generated and distributed across various nodes over networks. To preserve data privacy while extracting representative subsets, distributed coreset selection methods have received research attention \cite{hao2025fedcs,sivasubramanian2024gradient}. Nevertheless, current methods either neglect the robustness in coreset or remain confined to centralized settings. How to perform robust coreset selection within distributed networks \textit{remains under-explored}. Notably, existing robust coreset selection methods (e.g., \cite{dolatabadi2023adversarial,shinde2025data}) cannot be effectively applied in distributed robust coreset selection, as it is impractical to centralize distributed data due to data privacy, and performing selection locally \textit{fails to} fully exploit global information. Meanwhile, existing distributed coreset selection methods (e.g., \cite{hao2025fedcs,sivasubramanian2024gradient}) are also not directly applicable. This is because robust optimization introduces an additional optimization level into the original optimization problem, fundamentally nesting its structure. In fact, the distributed robust coreset selection problem admits \textit{no downward polynomial-time reduction} to its non-robust counterpart unless the polynomial hierarchy collapses (e.g., $\Sigma_r^{\mathrm P} = \Sigma_{r+1}^{\mathrm P}$) \cite{sugishita2026decision}, rendering it inherently intractable. To bridge this gap, this work takes a \textbf{pioneering step} towards addressing the distributed robust coreset selection problem.

\begin{table}[t]
\caption{Comparison in terms of key properties between the proposed F$^2$CTO and related state-of-the-art approaches.}
\renewcommand\arraystretch{1.1}
\renewcommand\tabcolsep{2.9pt}
\centering
\scalebox{1}{\begin{tabular}{l|ccccc}
\toprule
Method  & Bilevel  & Trilevel & First-order & Federated  &  Constrained
\\
\hline
 ADBO \cite{jiao2022asynchronous} &  \checkmark  &    &   \checkmark  &  \checkmark &    \\
 FEDNEST \cite{tarzanagh2022fednest} &  \checkmark  &    &   & \checkmark  &    \\
 BCSR \cite{hao2023bilevel} &  \checkmark  &    &   &   &  \checkmark  \\ 
 DIAMOND \cite{qiu2023diamond} &  \checkmark  &    &   &  \checkmark &  \checkmark  \\ 
 TSG \cite{giovannelli2025stochastic} &  \checkmark  &  \checkmark  &   &    &    \\
 DTZO \cite{jiaodtzo} &  \checkmark  &  \checkmark  &   & \checkmark &      \\
 AFTO \cite{jiao2024provably} &  \checkmark  &  \checkmark  & \checkmark  & \checkmark &      \\
\hline
\textbf{F$^2$CTO}   &  \checkmark &  \checkmark &  \checkmark  &  \checkmark  &  \checkmark \\
\bottomrule
\end{tabular}}
\label{table:1}
\end{table}

\subsection{Trilevel Optimization}
Trilevel optimization refers to a class of optimization problems characterized by a three-tiered nested structure, wherein the three levels are mutually interdependent and dynamically influence one another. Compared to bilevel optimization \cite{qiu2023diamond}, trilevel optimization is significantly more challenging to solve due to its highly complex nested structure. In fact, merely identifying a feasible solution within a trilevel framework necessitates solving an underlying bilevel optimization problem. Trilevel optimization has been widely used in machine learning (e.g., \cite{chien2026trilevel,gao2024enhancing,jiao2022timeautoad,jian2024tri}) and networking (e.g., \cite{ghorbani2020protection,yao2007trilevel}). Since data is often distributed across multiple nodes, distributed trilevel optimization has garnered attention as an effective approach to solving these problems without transmitting raw data. In \cite{jiao2024provably}, a federated trilevel optimization framework based on two-layer polyhedral approximation is introduced. A zeroth-order distributed trilevel optimization method is proposed to address the trilevel optimization problems without relying on gradients in \cite{jiaodtzo}. However, existing distributed trilevel optimization methods only consider the trilevel optimization without level-wise constraints. To the best of our knowledge, this is the \textbf{first work} that addresses the constrained trilevel optimization problems in a distributed manner and provides the non-asymptotic convergence guarantees. To explicitly illustrate the superiority of the proposed method, we provide a comprehensive comparison with existing state-of-the-art multilevel optimization methods regarding key properties and convergence rates (see Tables \ref{table:1} and \ref{table:2}).

\section{Distributed Robust Coreset Selection}
\begin{figure*}[t]
\centering\includegraphics[width=0.85\linewidth]{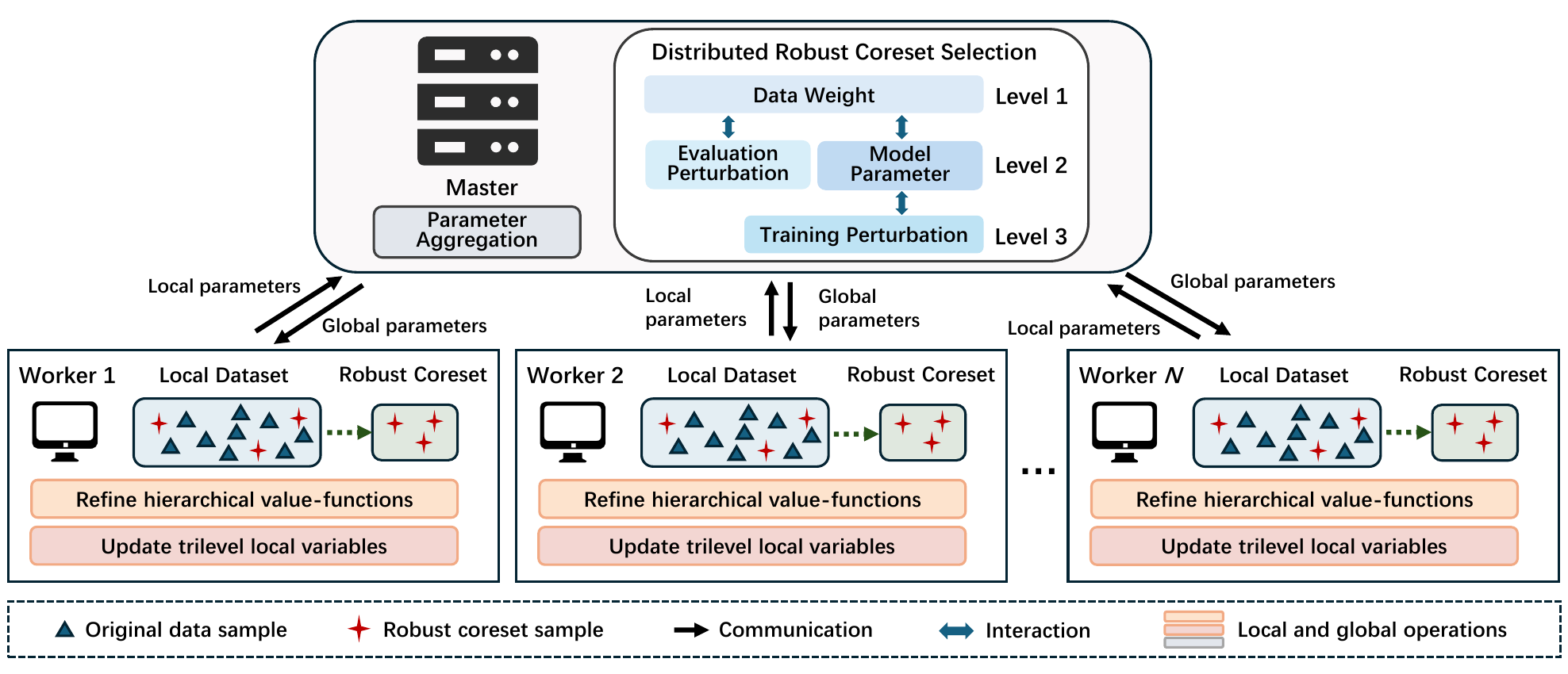}
\vspace{-2mm}
 \caption{Illustration of the proposed distributed robust coreset selection framework. The framework is formulated as a trilevel optimization problem. At the first level, the data weights ${\boldsymbol{\alpha}_i}$ are optimized to select the representative data samples. The second level serves two coupled purposes: generating the evaluation-side perturbations ${\boldsymbol{q}_i}$ for the first-level objective, and training a robust model $\boldsymbol{w}$ using the optimized data weights together with the training-side perturbations. The training-side perturbations ${\boldsymbol{p}_i}$ are obtained through the third-level optimization. During the distributed optimization, each worker locally refines hierarchical composite value-functions and updates variables, communicating only local parameters to the master without sharing raw data.}
    \label{fig:1}
\vspace{-2mm}
\end{figure*}

In this part, we first provide the motivating applications for investigating distributed robust coreset selection in Sec. \ref{sec:motivating}. Then, the fundamental interplay between coreset selection, robust optimization, and distributed learning is studied
in Sec. \ref{sec:3-A}, and distributed robust coreset selection is formulated as a trilevel optimization problem with level-wise constraints. To effectively solve it, a federated first-order constrained trilevel optimization method F$^2$CTO is proposed in Sec. \ref{sec.FFCTO}, which consists of a hierarchical composite value-function reformulation and a distributed alternating projected gradient algorithm.
The overview of the proposed framework is shown in Fig. \ref{fig:1}.

\subsection{Motivating Applications and Problem Setup}
\label{sec:motivating}

\subsubsection{Reliable Continual Learning}
Continual learning is well suited to IoT applications in which privacy-sensitive data streams are continuously generated at various edge devices \cite{chen2025online}. For example, in sensor-based human activity recognition, wearable devices continuously observe new activity patterns. Due to limited storage capacity, retaining all historical data is impractical, whereas learning only from new data may cause catastrophic forgetting \cite{schiemer2023online}. Furthermore, since adversarial perturbations to sensor data can mislead activity recognition models \cite{sah2019adar}, ensuring the reliability of continual learning is essential. This motivates our investigation on distributed robust coreset selection, which identifies robust coresets without sharing privacy-sensitive data, thereby reducing storage costs while preserving critical knowledge for reliable continual learning.

\subsubsection{Adversarially Robust Network Intrusion Detection}
Network intrusion detection aims to detect attacks and compromised devices in distributed IoT networks \cite{nguyen2019diot}. In such systems, massive volumes of privacy-sensitive traffic data are continuously generated and distributed across multiple nodes, training on all the data incurs substantial computation overhead \cite{zennaro2019analyzing}. Moreover, it is essential to defend against adversaries who deliberately manipulate network traffic characteristics to evade machine-learning-based intrusion detectors \cite{han2021evaluating}. To reduce training overhead while maintaining model robustness in network intrusion detection, we are motivated to study the distributed robust coreset selection for identifying representative coresets without transmitting local privacy-sensitive data.

\underline{Problem Setup.} Following the parameter-server architecture widely used in federated learning \cite{jiao2024provably}, we consider a distributed network comprising a master node (central server) and $N$ workers (clients). The global training dataset $\mathcal{D} = \bigcup_{i=1}^{N} \mathcal{D}_i$ is partitioned across the workers, where each local dataset $\mathcal{D}_i$ contains $M_i$ samples. Our objective is to construct a representative global coreset $\mathcal{D}^* = \bigcup_{i=1}^{N} \mathcal{D}_i^*$ (with $\mathcal{D}_i^* \subset \mathcal{D}_i$) that preserves the robust training objective of the full dataset while substantially reducing the sample size, i.e., $K_i \ll M_i$.

\begin{table}[t]
\caption{Comparison in terms of convergence rate between the proposed F$^2$CTO and related state-of-the-art approaches.}
\renewcommand\arraystretch{1.1}
\renewcommand\tabcolsep{3.6pt}
\centering
\scalebox{1}{\begin{tabular}{l|c|c|c|c}
\toprule
Method  & Bilevel$_{w/o}$  &  Bilevel$_{w}$  & Trilevel$_{w/o}$ & Trilevel$_{w}$
\\
\hline
 ADBO \cite{jiao2022asynchronous} & $\mathcal{O}(1/{\epsilon^2})$   &    &   &     \\
 FEDNEST \cite{tarzanagh2022fednest} &$\mathcal{O}(1/{\epsilon^2})$    &    &   &       \\
 BCSR \cite{hao2023bilevel} &    &  $\mathcal{O}(1/{\epsilon^2})$  &   &    \\
 DIAMOND \cite{qiu2023diamond} &    &  $\mathcal{O}(1/{\epsilon^{3/2}})$  &   & \\
 TSG \cite{giovannelli2025stochastic}  &    &    & $\mathcal{O}(1/{\epsilon^2})$  &      \\
 DTZO \cite{jiaodtzo} &    &    & $\mathcal{O}(1/{\epsilon^2})$  &      \\
 AFTO \cite{jiao2024provably} &    &    & $\mathcal{O}(1/{\epsilon^2})$  &      \\
\hline
\textbf{F$^2$CTO}  &    &    &   &   $\mathcal{O}(1/{\epsilon^{3/2}})$    \\
\bottomrule
\end{tabular}}
\label{table:2}
\begin{flushleft}
\footnotesize
Here, the subscripts $w$ and $w/o$ denote optimization with and without level-wise constraints, respectively.
\end{flushleft}
\vspace{-4mm}
\end{table}

\subsection{Overall Constrained Trilevel Optimization Problem}
\label{sec:3-A}

To mathematically formalize the distributed robust coreset selection, we adopt a multilevel optimization framework. Coreset selection is typically formulated as a bilevel optimization problem \cite{shen2026brief}. Building upon the continuous regularized bilevel framework proposed in \cite{hao2023bilevel}, we jointly incorporate min-max robust optimization and distributed learning. This allows us to cast the distributed robust coreset selection as the following trilevel optimization problem with level-wise constraints:
\vspace{-1mm}
\begin{equation}
\label{eq:key}
\begin{array}{l}
\min  \frac{1}{N}\sum\limits_{i = 1}^N {\sum\limits_{k = 1}^{{M_i}} {{\ell _{i,k}}(\boldsymbol{w};{x_{i,k}} \!+\! {q_{i,k}},{y_{i,k}})}  \!-\! \lambda \sum\limits_{b = 1}^{{K_i}}\! {{\mathbb{E}_{\boldsymbol{z}_i}}} {{({\boldsymbol{\alpha}_i} \!+\! \boldsymbol{z}_i)}_{[b]}}} \\
{\rm{s}}.{\rm{t}}.\; 0 \le {\alpha _{i,k}} \le 1,||{\boldsymbol{\alpha}_i}|| = 1, \forall i, k,
\\ \quad {\boldsymbol{q}_i} = \mathop {\arg \min }\limits_{{ {\boldsymbol{q}_i'}:||{q_{i,k}'}|{|_\infty } \le c_1, \forall k} }- \sum\limits_{k = 1}^{{M_i}}{\ell _{i,k}}(\boldsymbol{w};{x_{i,k}} \!+\! {q_{i,k}'},{y_{i,k}}), \forall i,\\
\quad \boldsymbol{w} = \arg {\min _{\boldsymbol{w}'}}\frac{1}{N}\sum\limits_{i = 1}^N {\sum\limits_{k = 1}^{{M_i}} {{\alpha _{i,k}}} {\ell _{i,k}}(\boldsymbol{w}';{x_{i,k}} \!+\! {p_{i,k}'},{y_{i,k}})} \\
\qquad{\rm{s}}.{\rm{t}}.\;||{\boldsymbol{w}'}|{|_2 } \le c_2,\\
\qquad {\boldsymbol{p}_i} = \arg {\min _{{\boldsymbol{p}_i'}}} - \sum\limits_{k = 1}^{{M_i}}\! {{\alpha _{i,k}}} {\ell _{i,k}}(\boldsymbol{w}';{x_{i,k}}\! +\! {p_{i,k}'},{y_{i,k}}),\forall i,\\
\qquad \quad {\rm{s}}.{\rm{t}}.\;||{p_{i,k}'}|{|_\infty } \le c_3, \forall k\\
{\rm{var}}.\qquad \qquad  \{\boldsymbol{\alpha}_i\}, \{{\boldsymbol{q}_i} \},\boldsymbol{w},\{\boldsymbol{p}_i\}.
\end{array}
\end{equation}
where $N$ represents the total number of workers. For each worker $i$, $M_i$ and $K_i$ denote the sizes of its local dataset and robust coreset, respectively. The pair $({x_{i,k}}, {y_{i,k}})$ denotes the $k^{\rm{th}}$ data sample and label, with ${\ell_{i,k}}(\cdot)$ being the corresponding loss function.  $\boldsymbol{\alpha}_i \in \mathbb{R}^{M_i}$ is the local data weight vector (where ${\alpha_{i,k}}$ is its $k^{\rm{th}}$ element), while $\boldsymbol{q}_i=[q_{i,1},\cdots,q_{i,M_i}]$ and $\boldsymbol{p}_i=[p_{i,1},\cdots,p_{i,M_i}]$ represent the evaluation-side and training-side adversarial perturbations, respectively. $\boldsymbol{w} \in \mathbb{R}^{d}$ denotes the model parameters, and $\boldsymbol{z}_i \sim \mathcal{N}(\mathbf{0}, \delta^2 \mathbf{I})$ is a Gaussian noise vector, where ${({\boldsymbol{\alpha}_i} + \boldsymbol{z}_i)}_{[b]}$ identifies the $b^{\rm{th}}$ largest component of the resulting noisy vector. Moreover, ${\alpha _{i,k}} \le 1,||{\boldsymbol{\alpha}_i}|| = 1, ||{q_{i,k}'}|{|_\infty } \le c_1, ||{\boldsymbol{w}'}|{|_2 } \le c_2, ||{p_{i,k}'}|{|_\infty } \le c_3, \forall i,k$ are the level-wise constraints, where $c_1 > 0, c_2 > 0, c_3 > 0$ are constants. To clearly articulate the overall trilevel optimization problem, we systematically analyze each level as follows:

\noindent\textbf{\ding{172} First-level optimization:} The first-level local objective ${{f_{1,i}}({\boldsymbol{\alpha}_i},{\boldsymbol{q}_i},\boldsymbol{w},{\boldsymbol{p}_i})}$ on worker $i$ consists of two parts. The first term $\sum\nolimits_{k = 1}^{{M_i}} {{\ell _{i,k}}(\boldsymbol{w};{x_{i,k}} + {q_{i,k}},{y_{i,k}})} $  evaluates the robust empirical loss of this trained model $\boldsymbol{w}$ on the full dataset under worst-case evaluation-side perturbations ${\boldsymbol{q}_i}$. Although $\boldsymbol{\alpha}_i$ does not appear explicitly in this term, it affects this term indirectly through the trained model $\boldsymbol{w}$. The second term $ - \lambda \sum\nolimits_{b = 1}^{{K_i}}\! {{\mathbb{E}_{\boldsymbol{z}_i}}} {{({\boldsymbol{\alpha}_i} \!+\! \boldsymbol{z}_i)}_{[b]}}$ acts as a smoothed top-$K_i$ regularization \cite{hao2023bilevel}, $\lambda>0$ is the regularization coefficient. Minimizing it encourages the largest $K_i$ entries of $\boldsymbol{\alpha}_i$ to carry more weight, thereby promoting a robust coreset of size $K_i$.

\noindent\textbf{\ding{173} Second-level optimization:} The second level consists of two optimization problems with different roles. The first one is associated with the evaluation-side perturbation $\boldsymbol{q}_i$ used in the first-level robust assessment, which is  ${{f_{2,i}^{(1)}}({\boldsymbol{\alpha}_i},{\boldsymbol{q}_i},\boldsymbol{w})}=- \sum\nolimits_{k = 1}^{{M_i}}{\ell _{i,k}}(\boldsymbol{w};{x_{i,k}} \!+\! {q_{i,k}},{y_{i,k}})$. This objective generates the worst-case evaluation perturbations for the first-level objective. The second one corresponds to the weighted robust training objective ${{f_{2,i}^{(2)}}({\boldsymbol{\alpha}_i},\boldsymbol{w},{\boldsymbol{p}_i})}=\sum\nolimits_{k = 1}^{{M_i}} {{\alpha _{i,k}}} {\ell _{i,k}}(\boldsymbol{w};{x_{i,k}} + {p_{i,k}},{y_{i,k}})$. This objective uses the data weights $\boldsymbol{\alpha}_i$ and the adversarial perturbations $\boldsymbol{p}_i$ determined by the first and third level to train the robust model. 

%Hence, the second level serves as a bridge between coreset selection and robust optimization.

\noindent\textbf{\ding{174} Third-level optimization:} The third level aims to generate the training-side perturbations for the second-level robust optimization. Given fixed data weights $\boldsymbol{\alpha}_i$ and candidate model parameters $\boldsymbol{w}'$,  the local third-level objective on worker $i$ is defined as ${f_{3,i}}({\boldsymbol{\alpha}_i},\boldsymbol{w}',{\boldsymbol{p}_i'})= - \sum\nolimits_{k = 1}^{{M_i}} {{\alpha _{i,k}}} {\ell _{i,k}}(\boldsymbol{w}';{x_{i,k}} + {p_{i,k}'},{y_{i,k}})$. Minimizing $f_{3,i}$ can effectively identify the worst-case perturbations for the second-level optimization. Since the loss is weighted by $\boldsymbol{\alpha}_i$, data assigned larger weights exert a stronger influence on the generated perturbations.

\noindent \underline{\textit{\textbf{Challenges in Solving the Problem (\ref{eq:key}):}}} In trilevel optimization, even identifying a feasible solution is NP-hard \cite{jiao2025pr}. From a polynomial hierarchy perspective, the decision versions of bilevel and trilevel linear optimization are $\Sigma_1^{\mathrm P}$-complete and $\Sigma_2^{\mathrm P}$-complete, respectively \cite{sugishita2026decision}. Unless polynomial hierarchy collapses (e.g., $\Sigma_r^{\mathrm P} \!=\! \Sigma_{r+1}^{\mathrm P}$), higher-level structures represent harder problem classes that admit no downward polynomial-time reduction. Moreover, unlike the distributed trilevel optimization studied in previous works, Eq. (\ref{eq:key}) presents \textit{two additional challenges:} 1) it incorporates auxiliary constraints at each level, and 2) it involves two distinct optimization problems at the second level. The complex interplay between these level-wise constraints and the multiple second-level subproblems significantly complicates the geometry in this optimization problem, \textit{making existing trilevel optimization methods fail to solve it, which motivates us to develop an efficient distributed constrained trilevel optimization method.}

\subsection{Federated First-order Constrained Trilevel Optimization}
\label{sec.FFCTO}
To effectively solve the trilevel problem (\ref{eq:key}), a federated first-order constrained trilevel optimization F$^2$CTO is proposed. Firstly, given that this highly nested structure is intractable to solve directly, Sec. \ref{sec:3-B} introduces a hierarchical composite value-function method, which transforms the problem into a single-level structure, making it amenable to first-order optimization algorithms. Building upon this, Sec. \ref{sec:3-C} presents a distributed first-order alternating projected gradient algorithm to efficiently address this problem.

\subsubsection{Hierarchical Composite Value-function Reformulation}
\label{sec:3-B}

Value-function methods are widely used in bilevel optimization to reduce the optimization problems into tractable structures \cite{yang2025first}. Motivated by this, this work proposes a hierarchical composite value-function reformulation tailored for constrained trilevel optimization, which consists of an inner-layer value-function and two outer-layer value-functions.  Specifically, the third-level problem, i.e., ${\boldsymbol{p}_i} = \arg\min_{{\boldsymbol{p}_i'}} {f_{3,i}}({\boldsymbol{\alpha}_i}, \boldsymbol{w}', {\boldsymbol{p}_i'})\,{\rm{s}}.{\rm{t}}.\,||{p_{i,k}'}|{|_\infty } \le c_3, \forall k$, operates as a constraint on the second-level optimization, which can be equivalently expressed by the value-function constraint:
\vspace{-1mm}
\begin{equation}
\label{eq:7_6_2}
\begin{array}{l}
{f_{3,i}}({\boldsymbol{\alpha}_i},\boldsymbol{w}',{\boldsymbol{p}_i}) - {V_{3,i}}({\boldsymbol{\alpha}_i},\boldsymbol{w}') \le 0,\forall i,\\
{V_{3,i}}({\boldsymbol{\alpha}_i},\boldsymbol{w}') = {\min _{\boldsymbol{p}_i':||{p_{i,k}'}|{|_\infty }  \le  c_3,\forall k}} {f_{3,i}}({\boldsymbol{\alpha}_i},\boldsymbol{w}',{\boldsymbol{p}_i}'),
\end{array}
\vspace{-1.25mm}
\end{equation}
where $V_{3,i}(\cdot)$ is the inner-layer value-function.
Thus, by utilizing the inner-layer value-function constraint in Eq. (\ref{eq:7_6_2}), the original optimization problem can be reformulated as a constrained bilevel optimization problem as follows:
\vspace{-1.25mm}
\begin{equation}
\label{eq:6_29_3}
\begin{array}{l}
\min  \frac{1}{N}\!\sum\limits_{i = 1}^N {\sum\limits_{k = 1}^{{M_i}} {{\ell _{i,k}}(\boldsymbol{w};{x_{i,k}} \!+\! {q_{i,k}},{y_{i,k}})} \! -\! \lambda \sum\limits_{b = 1}^{{K_i}} \! {{\mathbb{E}_{\boldsymbol{z}_i}}} {{({\boldsymbol{\alpha}_i} \!+\! \boldsymbol{z}_i)}_{[b]}}} \\
{\rm{s}}.{\rm{t}}.\; 0 \le {\alpha _{i,k}} \le 1,||{\boldsymbol{\alpha}_i}|| = 1,\forall i, k,
\\ \quad {\boldsymbol{q}_i} = \mathop {\arg \min }\limits_{{ {\boldsymbol{q}_i'}:||{q_{i,k}'}|{|_\infty } \le c_1, \forall k} }- \sum\limits_{k = 1}^{{M_i}}{\ell _{i,k}}(\boldsymbol{w};{x_{i,k}} \!+\! {q_{i,k}'},{y_{i,k}}), \forall i,\\
\quad \boldsymbol{w} = \arg {\min _{\boldsymbol{w}'}}\frac{1}{N}\sum\limits_{i = 1}^N {\sum\limits_{k = 1}^{{M_i}} {{\alpha _{i,k}}} {\ell _{i,k}}(\boldsymbol{w}';{x_{i,k}} \!+\! {p_{i,k}'},{y_{i,k}})} \\
\qquad{\rm{s}}.{\rm{t}}.\;||{\boldsymbol{w}'}|{|_2 } \!\le\! c_2, {f_{3,i}}({\boldsymbol{\alpha}_i},\boldsymbol{w}',{\boldsymbol{p}_i}) \!-\! {V_{3,i}}({\boldsymbol{\alpha}_i},\boldsymbol{w}') \! \le \! 0,\forall i,\\
{\rm{var}}.\qquad \qquad  \{\boldsymbol{\alpha}_i\}, \{{\boldsymbol{q}_i} \},\boldsymbol{w},\{\boldsymbol{p}_i\}.
\end{array}
\end{equation}

Likewise, in Eq. (\ref{eq:6_29_3}), the lower-level optimization also acts as a constraint on the upper-level problem. Specifically, among the two lower-level subproblems, the first can be equivalently expressed via the following value-function constraint:
\vspace{-1.25mm}
\begin{equation}
\label{eq:7_6_4}
\!\!\begin{array}{l}
{{f_{2,i}^{(1)}}({\boldsymbol{\alpha}_i},{\boldsymbol{q}_i},\boldsymbol{w})}\!-\!V_{2,i}^{(1)}({\boldsymbol{\alpha}_i},\boldsymbol{w}) \! \le \! 0,||{q_{i,k}}|{|_\infty } \!\le \! c_1, \forall i, k,\\
V_{2,i}^{(1)}({\boldsymbol{\alpha}_i},\boldsymbol{w})=\mathop {\min }\nolimits_{{ {\boldsymbol{q}_i'}:||{q_{i,k}'}|{|_\infty } \le c_1, \forall k} }{{f_{2,i}^{(1)}}({\boldsymbol{\alpha}_i},{\boldsymbol{q}_i'},\boldsymbol{w})},
\end{array}
\vspace{-1.25mm}
\end{equation}
where $V_{2,i}^{(1)}(\cdot)$ represents the first outer-layer value-function. Analogously, the second lower-level subproblem admits a similar reformulation, expressed as:
\vspace{-1.25mm}
\begin{equation}
\label{eq:7_6_5}
\begin{array}{l}
\frac{1}{N}\sum\nolimits_{i = 1}^N {{f_{2,i}^{(2)}}({\boldsymbol{\alpha}_i},\boldsymbol{w},{\boldsymbol{p}_i})}  - {V_2^{(2)}}(\{ {\boldsymbol{\alpha}_i}\} ) \le 0,\\
{V_2^{(2)}}(\{ {\boldsymbol{\alpha}_i}\} ) = {\min _{\boldsymbol{w}',\boldsymbol{p}_i'}}\sum\nolimits_{i = 1}^N {{f_{2,i}^{(2)}}({\boldsymbol{\alpha}_i},\boldsymbol{w}',{\boldsymbol{p}_i'})} \\
{\rm{s}}.{\rm{t}}.\;||{\boldsymbol{w}'}|{|_2 } \le c_2,||{p_{i,k}'}|{|_\infty } \le c_3, \forall i, k,\\
{f_{3,i}}({\boldsymbol{\alpha}_i},\boldsymbol{w}',{\boldsymbol{p}_i'}) - {V_{3,i}}({\boldsymbol{\alpha}_i},\boldsymbol{w}') \le 0,\forall i,
\end{array}
\vspace{-1.25mm}
\end{equation}
where ${V_2^{(2)}}(\cdot)$ denotes the second outer-layer value-function. 
Consequently, by incorporating these two outer-layer constraints in Eqs. (\ref{eq:7_6_4}) and (\ref{eq:7_6_5}), the resulting hierarchical composite optimization problem can be formulated as follows:
\vspace{-1.5mm}
\begin{equation}
\label{eq:7_6_6}
\begin{array}{l}
\min  \frac{1}{N}\!\sum\limits_{i = 1}^N \! {\sum\limits_{k = 1}^{{M_i}} \! {{\ell _{i,k}}(\boldsymbol{w};{x_{i,k}} \!+\! {q_{i,k}},{y_{i,k}})}  \!-\! \lambda \sum\limits_{b = 1}^{{K_i}} \! {{\mathbb{E}_{\boldsymbol{z}_i}}} {{({\boldsymbol{\alpha}_i} \!+\! \boldsymbol{z}_i)}_{[b]}}} \\
{\rm{s}}.{\rm{t}}.\;\;\; \; 0 \le {\alpha _{i,k}} \le 1,||{\boldsymbol{\alpha}_i}|| = 1,\forall i, k, ||{\boldsymbol{w}}|{|_2 } \!\le\! c_2,\\
\quad \;\;\; \;\,
||{q_{i,k}}|{|_\infty } \!\le\! c_1, ||{p_{i,k}}|{|_\infty } \!\le\! c_3,\forall i, k,
\\ \quad \;\;\; \;\, {{f_{2,i}^{(1)}}({\boldsymbol{\alpha}_i},{\boldsymbol{q}_i},\boldsymbol{w})}-V_{2,i}^{(1)}({\boldsymbol{\alpha}_i},\boldsymbol{w}) \le 0,\forall i,\\
\quad \;\;\;\;\,
\frac{1}{N}\sum\nolimits_{i = 1}^N {{f_{2,i}^{(2)}}({\boldsymbol{\alpha}_i},\boldsymbol{w},{\boldsymbol{p}_i})}  - {V_2^{(2)}}(\{ {\boldsymbol{\alpha}_i}\} ) \le 0,\\
\quad \;\;\;\;\,  {f_{3,i}}({\boldsymbol{\alpha}_i},\boldsymbol{w},{\boldsymbol{p}_i}) - {V_{3,i}}({\boldsymbol{\alpha}_i},\boldsymbol{w}) \le 0,\forall i,\\
{\rm{var}}.\qquad \qquad  \{\boldsymbol{\alpha}_i\}, \{{\boldsymbol{q}_i} \},\boldsymbol{w},\{\boldsymbol{p}_i\}.
\end{array}
\end{equation}

\subsubsection{Distributed Alternating Projected Gradient Algorithm}
\label{sec:3-C}

In this section, we propose a first-order distributed algorithm to effectively solve the resulting problem. The proposed method consists of two key steps: refining the hierarchical composite value-functions and updating the variables alternately. Specifically, in the $(t+1)^{\text{th}}$ iteration, inspired by the fact that 1) lower-level optimization often serves as a \textit{soft constraint} to the upper-level ones in multilevel optimization \cite{jiao2025pr}, which can be approximated to a certain extent without resulting in meaningless solutions, and 2) it is common practice to employ multiple gradient descent steps to approximate the optimal lower-level solution \cite{yang2025first}. Each worker $i$ performs $R$ steps of projected gradient descent to update value-functions ${V_{3,i}}({\boldsymbol{\alpha}_i^t},{\boldsymbol{w}^t})={f_{3,i}}(\boldsymbol{\alpha}_i^t,{\boldsymbol{w}^t},{\boldsymbol{p}_i^*}),{\boldsymbol{p}_i^*} = \arg {\min _{{\boldsymbol{p}_i'} \in \boldsymbol{P}}}{f_{3,i}}(\boldsymbol{\alpha}_i^t,{\boldsymbol{w}^t},{\boldsymbol{p}_i'})$ and $V_{2,i}^{(1)}({\boldsymbol{\alpha}_i^t},\boldsymbol{w}^t)={{f_{2,i}^{(1)}}({\boldsymbol{\alpha}_i^t},{\boldsymbol{q}_i^*},\boldsymbol{w}^t)}$, $\boldsymbol{q}_i^*=\mathop  {\arg\min }\nolimits_{{ {\boldsymbol{q}_i'}\in \boldsymbol{Q}} }{{f_{2,i}^{(1)}}({\boldsymbol{\alpha}_i^t},{\boldsymbol{q}_i'},\boldsymbol{w}^t)}$ as:
\vspace{-1.25mm}
\begin{equation}
\label{eq:7_9_7}
\boldsymbol{p}_i^{r+1} = {\mathcal{P}_{\boldsymbol{P}}}(\boldsymbol{p}_i^{r} - \eta_{\boldsymbol{p}} \nabla_{\boldsymbol{p}} {f_{3,i}}(\boldsymbol{\alpha}_i^t,{\boldsymbol{w}^t},{\boldsymbol{p}_i^r})), {\boldsymbol{p}_i^*} \approx \boldsymbol{p}_i^{R},
\end{equation}
\vspace{-5mm}
\begin{equation}
\boldsymbol{q}_i^{r+1} = {\mathcal{P}_{\boldsymbol{Q}}}(\boldsymbol{q}_i^{r} - \eta_{\boldsymbol{q}} \nabla_{\boldsymbol{q}} {f_{2,i}^{(1)}}({\boldsymbol{\alpha}_i^t},{\boldsymbol{q}_i^{r}},\boldsymbol{w}^t), {\boldsymbol{q}_i^*} \approx \boldsymbol{q}_i^{R},
\vspace{-1.25mm}
\end{equation}
where ${\mathcal{P}_{\boldsymbol{P}}}$ and ${\mathcal{P}_{\boldsymbol{Q}}}$ are the projection operator onto the feasible sets ${\boldsymbol{Q}}=\{{\boldsymbol{q}_i}:||{q_{i,k}}|{|_\infty } \le c_1, \forall k\}$ and ${\boldsymbol{P}}=\{{\boldsymbol{p}_i}:||{p_{i,k}}|{|_\infty } \le c_3, \forall k\}$, respectively. $\eta_{\boldsymbol{p}}$ and $\eta_{\boldsymbol{q}}$ are step-sizes. Likewise, for the second outer-layer value-function $V_2^{(2)}(\{\boldsymbol{\alpha}_i^t\}) = \sum_{i} f_{2,i}(\boldsymbol{\alpha}_i^t, \boldsymbol{w}^*, \boldsymbol{p}_i^*)$, where $\boldsymbol{w}^*, \boldsymbol{p}_i^* = \arg \min_{\boldsymbol{w}' \in \boldsymbol{W}, \boldsymbol{p}_i' \in \boldsymbol{P}} \sum_{i} f_{2,i}(\boldsymbol{\alpha}_i^t, \boldsymbol{w}', \boldsymbol{p}_i') + \frac{\phi}{N} \big( \sum_{i} f_{3,i}(\boldsymbol{\alpha}_i^t, \boldsymbol{w}', \boldsymbol{p}_i') - V_{3,i}(\boldsymbol{\alpha}_i^t, \boldsymbol{w}') \big)$, we approximate the optimal $\boldsymbol{w}^*$ and $\boldsymbol{p}_i^*$ using the results after $\hat{R}$ steps of alternating projected gradient descent. Specifically, defining $g_{2,i}(\boldsymbol{\alpha}_i, \boldsymbol{w}, \boldsymbol{p}_i) = f_{2,i}^{(2)}(\boldsymbol{\alpha}_i, \boldsymbol{w}, \boldsymbol{p}_i) + \phi \big( f_{3,i}(\boldsymbol{\alpha}_i, \boldsymbol{w}, \boldsymbol{p}_i) - V_{3,i}(\boldsymbol{\alpha}_i, \boldsymbol{w}) \big)$, where $\phi$ is a penalty parameter, for each $r = 0, \cdots, \hat{R}-1$, worker $i$ computes:
\vspace{-1.25mm}
\begin{equation}
\begin{array}{l}
\boldsymbol{p}_i^{r + 1} = {{\mathcal{P}_{\boldsymbol{P}}}}(\boldsymbol{p}_i^r - {\eta _{\boldsymbol{p}}}{\nabla _{\boldsymbol{p}}}{g_{2,i}}(\boldsymbol{\alpha}_i^t,{\boldsymbol{w}_i^{r}},\boldsymbol{p}_i^r)),\\
\boldsymbol{w}_i^{r + 1} = {{\mathcal{P}_{\boldsymbol{W}}}}({\boldsymbol{w}_i^{r}} - {\eta _{\boldsymbol{w}}}{\nabla _{\boldsymbol{w}}}{g_{2,i}}(\boldsymbol{\alpha}_i^t,{\boldsymbol{w}_i^{r}},\boldsymbol{p}_i^{r + 1})),
\end{array}
\vspace{-1.25mm}
\end{equation}
where ${{\mathcal{P}_{\boldsymbol{W}}}}$ is the projection operator onto the feasible set ${\boldsymbol{W}}=\{{\boldsymbol{w}}:||{\boldsymbol{w}}|{|_2 } \le c_2\}$, and ${\eta _{\boldsymbol{w}}}$ is the step-size.
Then, the local parameters $\boldsymbol{w}_i^{\hat{R}}$ are transmitted to the master, which computes their average as $\boldsymbol{w}^{\hat{R}} = \frac{1}{N}\sum_{i=1}^N \boldsymbol{w}_i^{\hat{R}}$. Thus, the approximations can be obtained: $\boldsymbol{w}^* \approx \boldsymbol{w}^{\hat{R}}$ and $\boldsymbol{p}_i^* \approx \boldsymbol{p}_i^{\hat{R}}$.

Building upon the refined hierarchical composite value-functions and incorporating the penalty method commonly used in bilevel optimization \cite{nazari2025penalty}, we formulate the penalty function for the problem in Eq. (\ref{eq:7_6_6}) as follows:
\vspace{-1.25mm}
\begin{equation}
\label{eq:7_6_10}
\begin{array}{l}
\min \mathcal{L}(\{ {\boldsymbol{\alpha}_i}\},\{ {\boldsymbol{q}_i}\},\boldsymbol{w},\{ {\boldsymbol{p}_i}\} ) = \frac{1}{N} \! \sum\nolimits_{i = 1}^N \! {{L_{i}}({\boldsymbol{\alpha}_i},{\boldsymbol{q}_i},\boldsymbol{w},{\boldsymbol{p}_i})} \\ \! = \!\! \frac{1}{N}\!\!\sum\limits_{i = 1}^N \! {{f_{1,i}}({\boldsymbol{\alpha}_i},{\boldsymbol{q}_i},\boldsymbol{w},{\boldsymbol{p}_i})}  
\!+\!\!\frac{{{\rho_1}}}{N}\!\!\sum\limits_{i = 1}^N\!{{f_{2,i}^{(1)}}\!({\boldsymbol{\alpha}_i},{\boldsymbol{q}_i},\boldsymbol{w})}\!-\!V_{2,i}^{(1)}\!({\boldsymbol{\alpha}_i},\!\boldsymbol{w})\\
\;\; + \frac{{{\rho_2}}}{N}(\sum\nolimits_{i = 1}^ N  \! {{f_{2,i}^{(2)}}({\boldsymbol{\alpha}_i},\boldsymbol{w},{\boldsymbol{p}_i})} \! - \! {V_2^{(2)}}(  \{ {\boldsymbol{\alpha}_i}  \} ))
\\ \;\; + \frac{{{\rho_3}}}{N}\!\sum\nolimits_{i = 1}^N \! {{f_{3,i}}({\boldsymbol{\alpha}_i},\boldsymbol{w},{\boldsymbol{p}_i})\! -\! {V_{3,i}}({\boldsymbol{\alpha}_i},\boldsymbol{w})} \\
{\rm{s}}.{\rm{t}}. \; \; 0 \le {\alpha _{i,k}} \le 1,||{\boldsymbol{\alpha}_i}|| = 1,\forall i, k,||{\boldsymbol{w}}|{|_2 } \!\le\! c_2,
\\
\qquad   ||{q_{i,k}}|{|_\infty } \!\le\! c_1, ||{p_{i,k}}|{|_\infty } \!\le\! c_3,\forall i, k,
\\ 
{\rm{var}}.\qquad \qquad  \{\boldsymbol{\alpha}_i\}, \{{\boldsymbol{q}_i} \},\boldsymbol{w},\{\boldsymbol{p}_i\},
\end{array}
\end{equation}
where $\rho_1,\rho_2,\rho_3$ are the penalty parameters. To optimize the penalized objective in Eq. (\ref{eq:7_6_10}), and owing to the variable dependencies across the trilevel architecture, the local variables on worker $i$ are updated in a \textit{Gauss-Seidel} manner as follows:
\vspace{-2mm}
\begin{equation}
\label{eq:7_2_11}
\begin{array}{l}
\boldsymbol{\alpha}_i^{t + 1} = {\mathcal{P}_{\boldsymbol{A}}}({\boldsymbol{\alpha}_i ^t} - {\eta _{\boldsymbol{\alpha}} }{\nabla _{\boldsymbol{\alpha}} }\mathcal{L}_i(\boldsymbol{\alpha}_i^t,\boldsymbol{q}_i^t,{{\boldsymbol{w}^t}},\boldsymbol{p}_i^t)),
\end{array}
\vspace{-2mm}
\end{equation}
\begin{equation}
\label{eq:7_6_12}
\begin{array}{l}
\boldsymbol{q}_i^{t + 1} = {\mathcal{P}_{\boldsymbol{Q}}}({\boldsymbol{q}_i^t} - {\eta _{\boldsymbol{q}} }{\nabla _{\boldsymbol{q}} }\mathcal{L}_i(\boldsymbol{\alpha}_i^{t+1},\boldsymbol{q}_i^t,{{\boldsymbol{w}^t}},\boldsymbol{p}_i^t)),
\end{array}
\vspace{-2mm}
\end{equation}
\begin{equation}
\begin{array}{l}
\boldsymbol{w}_i^{t + 1} = {{\boldsymbol{w}^t}} - {\eta _{\boldsymbol{w}}}{\nabla _{\boldsymbol{w}}}\mathcal{L}_i(\boldsymbol{\alpha}_i^{t + 1},\boldsymbol{q}_i^{t+1},{{\boldsymbol{w}^t}},\boldsymbol{p}_i^t),
\end{array}
\vspace{-2mm}
\end{equation}
\begin{equation}
\label{eq:7_6_14}
\begin{array}{l}
\boldsymbol{p}_i^{t + 1} = {{\mathcal{P}_{\boldsymbol{P}}}}(\boldsymbol{p}_i^t - {\eta _{\boldsymbol{p}}}{\nabla _{\boldsymbol{p}}}\mathcal{L}_i(\boldsymbol{\alpha}_i^{t + 1},\boldsymbol{q}_i^{t+1},{{\mathcal{P}_{\boldsymbol{W}}}}({\boldsymbol{w}}_i^{t + 1}),\boldsymbol{p}_i^t)), 
\end{array}
\end{equation}
where ${\mathcal{P}_{\boldsymbol{A}}}$ represents the projection operator onto the feasible set ${\boldsymbol{A}}=\{{\boldsymbol{\alpha}_i}: 0 \le {\alpha _{i,k}} \le 1, \forall k,||{\boldsymbol{\alpha}_i}|| = 1\}$, and $\eta _{\boldsymbol{\alpha}}$ is the step-size.
Subsequently, the updated local model parameters $\boldsymbol{w}_i^{t + 1}$ are transmitted to the master, which then updates and broadcasts the global model parameters as follows:
\vspace{-1mm}
\begin{equation}
\label{eq:7_2_15}
\begin{array}{l}
{{\boldsymbol{w}^{t + 1}}} = {{\mathcal{P}_{\boldsymbol{W}}}}\left( {\frac{1}{N}\sum\nolimits_{i = 1}^N {\boldsymbol{w}_i^{t + 1}} } \right).
\end{array}
\end{equation}

It can be seen from Eq. (\ref{eq:7_9_7}) to Eq. (\ref{eq:7_2_15}) that the projection operation at each step is \textit{computationally efficient} due to the highly structured nature of the closed convex sets ${\boldsymbol{A}}, {\boldsymbol{Q}}, {\boldsymbol{W}}, {\boldsymbol{P}}$ \cite{bertsekas1997nonlinear}. Furthermore, we emphasize that the proposed method is a single-loop \textit{first-order} distributed algorithm that avoids the computation of hyper-gradients. The complete procedure of the proposed method is summarized in Algorithm \ref{algorithm}.

\section{Theoretical Analyses}

\begin{definition}
\label{def:1}
\textbf{(Stationarity gap)} Following \cite{lan2024projected,jiao2024provably,jiao2022distributed}, the projected gradient mappings at the $t^{\rm{th}}$ iteration are defined as $g_{\boldsymbol{\alpha} ,i}^t = \frac{1}{{{\eta _{\boldsymbol{\alpha}} }}N}(\boldsymbol{\alpha}_i^t - \boldsymbol{\alpha}_i^{t + 1})$, $g_{\boldsymbol{q} ,i}^t = \frac{1}{{{\eta _{\boldsymbol{q}} }}N}(\boldsymbol{q}_i^t - \boldsymbol{q}_i^{t + 1})$, $g_{\boldsymbol{w}}^t = \frac{1}{{{\eta _{\boldsymbol{w}}}}}({{\boldsymbol{w}^t}} - {{\boldsymbol{w}^{t + 1}}})$, and $g_{\boldsymbol{p},i}^t = \frac{1}{{{\eta _{\boldsymbol{p}}}}N}(\boldsymbol{p}_i^t - \boldsymbol{p}_i^{t + 1})$ for $i\!=\!1,\!\cdots\!,N$. Consequently, the stationarity gap of the studied problem at the $t^{\rm{th}}$ iteration can be expressed as:
\vspace{-1mm}
\begin{equation}
{G^t} = \left[ \begin{array}{l}
\frac{1}{{{\eta _{\boldsymbol{\alpha}} }}N}(\boldsymbol{\alpha}_i^t - \boldsymbol{\alpha}_i^{t + 1}),\forall i\\
\frac{1}{{{\eta _{\boldsymbol{q}} }}N}(\boldsymbol{q}_i^t - \boldsymbol{q}_i^{t + 1}), \forall i\\
\frac{1}{{{\eta _{\boldsymbol{w}}}}}({{\boldsymbol{w}^t}} - {{\boldsymbol{w}^{t + 1}}})\\
\frac{1}{{{\eta _{\boldsymbol{p}}}}N}(\boldsymbol{p}_i^t - \boldsymbol{p}_i^{t + 1}),\forall i
\end{array} \right].
\vspace{-1mm}
\end{equation}

Thus, we can also obtain: $||{G^t}|{|^2} = \sum\nolimits_{i = 1}^N {||g_{\boldsymbol{\alpha} ,i}^t|{|^2}} + \sum\nolimits_{i = 1}^N {||g_{\boldsymbol{q} ,i}^t|{|^2}}  + ||g_{\boldsymbol{w}}^t|{|^2} + \sum\nolimits_{i = 1}^N {||g_{\boldsymbol{p},i}^t|{|^2}} .$

\end{definition}

\begin{definition}
\textbf{($\epsilon$-stationary point)}
\label{def:2}
$(\{ {\boldsymbol{\alpha}_i^t}\},\{ {\boldsymbol{q}_i^t}\},\boldsymbol{w}^t,\{ {\boldsymbol{p}_i^t}\} )$ is defined as an $\epsilon$-stationary point when $||G^t||^2 \le \epsilon$, and $T(\epsilon)$ is defined as the first iteration to achieve the $\epsilon$-stationary point, i.e., $T(\epsilon)=\min \{t| \; ||G^t||^2 \le \epsilon \}$.
\end{definition}

\begin{algorithm}[t]
   \caption{F$^2$CTO: \underline{F}ederated \underline{F}irst-order \underline{C}onstrained \underline{T}rilevel \underline{O}ptimization}
\begin{algorithmic}
   \STATE {\bfseries Initialization:}  iteration $t = 0$, variables $\{ {\boldsymbol{\alpha}_i^0}\}$, $\{ {\boldsymbol{q}_i^0}\}$, $\boldsymbol{w}^0$, $\{ {\boldsymbol{p}_i^0}\}$, $i=1,\cdots,N$.
   
   \REPEAT

   \STATE \textit{\underline{Refinement of Hierarchical Composite Value-functions:}}
   \FOR{\emph{local worker $i$}}
   \STATE computes $\boldsymbol{p}_i^{R}$ to update value-function ${V_{3,i}}({\boldsymbol{\alpha}_i^t},{\boldsymbol{w}^t})$;

   \STATE computes $\boldsymbol{q}_i^{R}$ to update value-function $V_{2,i}^{(1)}({\boldsymbol{\alpha}_i^t},\boldsymbol{w}^t)$;

   \STATE computes ${\boldsymbol{w}}_i^{\hat{R}}, \boldsymbol{p}_i^{\hat{R}}$ and transmits ${\boldsymbol{w}}_i^{\hat{R}}$ to the master.
   \ENDFOR
   
   \FOR{\emph{master}}
   \STATE aggregates local parameters as ${\boldsymbol{w}}^{\hat{R}}=\frac{1}{N}\sum_{i=1}^N {\boldsymbol{w}}_i^{\hat{R}}$;
   \STATE broadcasts ${\boldsymbol{w}}^{\hat{R}}$ to update value-function ${V_2^{(2)}}(\boldsymbol{\alpha}_i^t)$.

   \ENDFOR

   \STATE \textit{\underline{Update of Optimization Variables:}}
   \FOR{\emph{local worker $i$}}
   \STATE updates local variables $\boldsymbol{\alpha}_i^{t + 1},\boldsymbol{q}_i^{t + 1},\boldsymbol{w}_i^{t + 1},\boldsymbol{p}_i^{t + 1}$ in a Gauss-Seidel manner according to Eq. (\ref{eq:7_2_11})-Eq. (\ref{eq:7_6_14}):
  
   \STATE transmits the updated $\boldsymbol{w}_i^{t + 1}$ to the master.

   \ENDFOR
   
   \FOR{\emph{master}}
   \STATE aggregates local model parameters to get the global model parameters ${{\boldsymbol{w}^{t + 1}}} = {{\mathcal{P}_{\boldsymbol{W}}}}( {\frac{1}{N}\sum\nolimits_{i = 1}^N {\boldsymbol{w}_i^{t + 1}} })$;
   
   \STATE broadcasts the obtained ${{\boldsymbol{w}^{t + 1}}}$ to the workers.

   \ENDFOR

   \STATE $t =t+1$;
   %\UNTIL{convergence}
   \UNTIL{termination.}

\end{algorithmic}
\label{algorithm}
\end{algorithm}

\begin{assumption}
\label{assum:1}
\textbf{($L$-smoothness)} Following previous work \cite{hao2023bilevel,jiao2024provably,liu2022bome}, we assume  function $\mathcal{L}_i(\cdot)$ has an $L$-Lipschitz continuous gradient. That is, there exists a constant $0<L<\infty$ such that for any $\boldsymbol{u} = (\boldsymbol{\alpha}_i, \boldsymbol{q}_i, \boldsymbol{w}, \boldsymbol{p}_i)$ and $\boldsymbol{u}' = (\boldsymbol{\alpha}_i', \boldsymbol{q}_i', \boldsymbol{w}', \boldsymbol{p}_i')$, we have:
\vspace{-1mm}
\begin{equation}
\label{eq:7_2_18}
||\nabla \mathcal{L}_i(\boldsymbol{u})- \nabla\mathcal{L}_i(\boldsymbol{u}')|| \le L||\boldsymbol{u}-\boldsymbol{u}'||.
\end{equation}

\vspace{-1mm}

\begin{table*}[t]
\renewcommand\arraystretch{1.3}
\renewcommand\tabcolsep{4.5pt}
\centering
\caption{Comparison of the proposed F$^2$CTO with state-of-the-art methods in terms of average robustness ($\%$) against various adversarial attacks across multiple datasets, all experiments were repeated five times.}
\label{tab:experiment1}
\begin{tabular}{l|ccc|ccc|ccc}
\toprule
\multirow{2}{*}{Methods}
& \multicolumn{3}{c|}{Permuted MNIST}
& \multicolumn{3}{c|}{Split CIFAR-100}
& \multicolumn{3}{c}{Tiny-ImageNet} \\
& FGSM & PGD & AutoAttack
& FGSM & PGD & AutoAttack
& FGSM & PGD & AutoAttack \\
\hline

BCSR \cite{hao2023bilevel}
& 50.12\text{$\pm$0.58}
& 46.25\text{$\pm$0.87}
& 42.93\text{$\pm$1.15}
& 40.22\text{$\pm$0.78}
& 38.04\text{$\pm$0.81}
& 36.55\text{$\pm$0.94}
& 29.93\text{$\pm$0.80}
& 28.30\text{$\pm$0.83}
& 26.41\text{$\pm$0.94} \\

Greedy Coreset \cite{borsos2020coresets}
& 50.46\text{$\pm$1.30}
& 47.00\text{$\pm$0.73}
& 45.13\text{$\pm$1.04}
& 45.76\text{$\pm$0.43}
& 43.79\text{$\pm$0.25}
& 42.66\text{$\pm$0.30}
& 36.91\text{$\pm$0.35}
& 35.25\text{$\pm$0.42}
& 33.59\text{$\pm$0.42} \\

FedCS \cite{hao2025fedcs}
& 49.25\text{$\pm$0.21}
& 45.17\text{$\pm$0.86}
& 41.97\text{$\pm$0.89}
& 44.47\text{$\pm$0.43}
& 41.97\text{$\pm$0.36}
& 40.62\text{$\pm$0.59}
& 34.02\text{$\pm$0.51}
& 32.29\text{$\pm$0.53}
& 30.58\text{$\pm$0.58} \\

GCFL \cite{sivasubramanian2024gradient}
& 49.08\text{$\pm$0.54}
& 45.24\text{$\pm$0.92}
& 42.54\text{$\pm$0.95}
& 43.29\text{$\pm$0.68}
& 41.73\text{$\pm$0.51}
& 40.88\text{$\pm$0.62}
& 34.16\text{$\pm$0.49}
& 32.65\text{$\pm$0.52}
& 31.62\text{$\pm$0.41} \\

ACS \cite{dolatabadi2023adversarial}
& 50.46\text{$\pm$0.76}
& 45.53\text{$\pm$0.86}
& 42.32\text{$\pm$1.36}
& 45.42\text{$\pm$0.31}
& 43.53\text{$\pm$0.26}
& 41.20\text{$\pm$0.62}
& 35.93\text{$\pm$0.37}
& 33.94\text{$\pm$0.55}
& 32.75\text{$\pm$0.52} \\

DTZO \cite{jiaodtzo}
& 49.62\text{$\pm$1.08}
& 45.58\text{$\pm$1.12}
& 41.91\text{$\pm$1.29}
& 44.32\text{$\pm$0.99}
& 42.28\text{$\pm$1.09}
& 40.99\text{$\pm$0.96}
& 33.60\text{$\pm$0.62}
& 31.72\text{$\pm$0.71}
& 30.07\text{$\pm$0.84} \\

AFTO \cite{jiao2024provably}
& 50.23\text{$\pm$0.58}
& 46.16\text{$\pm$0.41}
& 43.33\text{$\pm$0.64}
& 45.81\text{$\pm$0.89}
& 43.75\text{$\pm$0.69}
& 42.58\text{$\pm$0.74}
& 35.65\text{$\pm$0.58}
& 33.51\text{$\pm$0.59}
& 32.22\text{$\pm$0.64} \\

\hline
\textbf{F$^2$CTO}
& \textbf{54.21\text{$\pm$0.40}}
& \textbf{51.10\text{$\pm$0.51}}
& \textbf{48.56\text{$\pm$0.96}}
& \textbf{48.22\text{$\pm$0.62}}
& \textbf{46.32\text{$\pm$0.85}}
& \textbf{45.84\text{$\pm$0.73}}
& \textbf{40.99\text{$\pm$1.64}}
& \textbf{39.32\text{$\pm$1.57}}
& \textbf{38.18\text{$\pm$1.73}} \\
\bottomrule
\end{tabular}
\vspace{-2mm}
\end{table*}

By combining Eq. (\ref{eq:7_2_18}) with the Triangle inequality, we can derive that function $\mathcal{L}(\cdot)$ also satisfies the $L$-smoothness:
\vspace{-1mm}
\begin{equation}
\label{eq:7_2_19}
\begin{array}{l}
||\nabla \mathcal{L}(\boldsymbol{u})\!-\! \nabla\mathcal{L}(\boldsymbol{u}')|| \!= \!|| \frac{1}{N}\! \sum_{i=1}^N \! \nabla \mathcal{L}_i(\boldsymbol{u}) \!-\! \nabla\mathcal{L}_i(\boldsymbol{u}') ||
\\ \qquad \qquad \qquad  \qquad \;  \le L||\boldsymbol{u}-\boldsymbol{u}'||.
\end{array}
\end{equation}

\end{assumption}

\vspace{-1mm}

% Before proving the convergence results, we first present two auxiliary lemmas. The first lemma controls the discrepancy between the local projected model and the aggregated global model, while the second lemma establishes the one-step descent inequality for the full update.

\begin{lemma} 
\label{lemma:1}\textbf{(Local-Global Projection Gap)}
Under Assumption 1, the discrepancy between the local projected update ${{\mathcal{P}_{\boldsymbol{W}}}}({\boldsymbol{w}_i^{t+1}})$ and the aggregated global update ${{\boldsymbol{w}^{t + 1}}}$
is uniformly bounded. Specifically, for any worker $i$, we have:
\vspace{-1mm}
\begin{equation}
||{{\mathcal{P}_{\boldsymbol{W}}}}({\boldsymbol{w}_i^{t+1}}) - {{\boldsymbol{w}^{t + 1}}}||  \le 2{\eta _{\boldsymbol{w}}}{B_{\boldsymbol{w}}},\vspace{-1.25mm}
\end{equation}
where $ {B_{\boldsymbol{w}}} = \max_{i} \max_{({\boldsymbol{\alpha}_i}, {\boldsymbol{q}_i}, \boldsymbol{w}, {\boldsymbol{p}_i})} ||{\nabla _{\boldsymbol{w}}}\mathcal{L}_i({\boldsymbol{\alpha}_i}, {\boldsymbol{q}_i}, \boldsymbol{w}, {\boldsymbol{p}_i})||  $ is a finite constant due to the smoothness of $\mathcal{L}_i$ over the compact feasible domains, and ${\eta _{\boldsymbol{w}}}$ is the step-size.
\end{lemma}

\begin{lemma}
\label{lemma:2}
\textbf{(One-Step Descent Inequality)} Under Assumption 1, one complete iteration of the proposed algorithm decreases the objective function up to a controllable error. Specifically, the following inequality holds:
\vspace{-1mm}
\begin{equation}
\label{eq:lemma2}
\begin{array}{l}
\mathcal{L}(\{ \boldsymbol{\alpha}_i^{t + 1}\} ,\! \{\boldsymbol{q}_i^{t + 1}\},{{\boldsymbol{w}^{t + 1}}},\!\{ \boldsymbol{p}_i^{t + 1}\} ) \!-\! \mathcal{L}(\{ \boldsymbol{\alpha}_i^{t}\},\!\{\boldsymbol{q}_i^{t}\},{\boldsymbol{w}^{t}},\!\{ \boldsymbol{p}_i^t\} )\\
\le- \frac{1}{N}\sum\nolimits_{i = 1}^N ( {\frac{1}{{{\eta _{\boldsymbol{\alpha}} }}} - \frac{L}{2}} )||{\boldsymbol{\alpha}_i ^{t + 1}} - {\boldsymbol{\alpha}_i ^t}|{|^2}\\
-( {\frac{1}{{{\eta _{\boldsymbol{w}} }}} - \frac{L}{2}} )||{{\boldsymbol{w}^{t+1}}} - {{\boldsymbol{w}^t}}|{|^2} +
\frac{{2{L^2}{\eta _{\boldsymbol{w}}^2}{B_{\boldsymbol{w}}^2}}} {{\frac{1}{{{\eta _{\boldsymbol{p}}}}} - \frac{L}{2}}} \\
- \frac{1}{2N}\! \sum_{i = 1}^N\! ( (\frac{1}{\eta_{\boldsymbol{p}}} \!-\! \frac{L}{2}) ||\boldsymbol{p}_i^{t + 1} \!-\! \boldsymbol{p}_i^t||^2 \!+\! (\frac{1}{\eta_{\boldsymbol{q}}} \!-\! \frac{L}{2}) ||\boldsymbol{q}_i^{t + 1} \!- \!\boldsymbol{q}_i^t||^2 ).
\end{array}
\end{equation}
\end{lemma}

\begin{theorem}
\label{theorem:1}
\textbf{(Iteration Complexity)} Under Assumption \ref{assum:1}, setting step-sizes ${\eta _{\boldsymbol{\alpha}} }={\eta _{\boldsymbol{q}}} = {\eta _{\boldsymbol{w}}} = {\eta _{\boldsymbol{p}}} = \eta  = {T^{ - 1/3}}$, when $T\ge L^3$, the iteration complexity to achieve $\epsilon$-stationary point is upper bounded by
\vspace{-1mm}
\begin{equation}
\begin{array}{l}
 T(\epsilon) \! \sim  \! \max \left\{ {{L^3},\frac{{{{( {8( {\mathcal{L}(\{\boldsymbol{\alpha}_i ^0\},\{\boldsymbol{q}_i^0\},{\boldsymbol{w}^0},\{\boldsymbol{p}_i^0\}) - \mathcal{L}^*} ) + 32{L^2}{B_{\boldsymbol{w}}^2}} )}^{3/2}}}}{{{\epsilon^{3/2}}}}} \right\} \! ,
\end{array}
\end{equation}
\end{theorem}
where $\mathcal{L}^*=\min\mathcal{L}(\{\boldsymbol{\alpha}_i \},\{\boldsymbol{q}_i\},{\boldsymbol{w}},\{\boldsymbol{p}_i\})$ is a constant due to the continuity of $\mathcal{L}$ over the compact sets, and $ {B_{\boldsymbol{w}}}$ is also a constant. The detailed proofs are shown in Sec. \ref{appendix:1}.

\begin{theorem}
\label{theorem:2}
\textbf{(Communication Complexity)} Following \cite{jiao2024provably}, communication complexity is defined as the total volume of information transmitted until the algorithm converges. The communication complexity for the proposed method to achieve the $\epsilon$-stationary point is $C_{\mathrm{comm}}(\epsilon)=\mathcal{O}(\frac{d}{{\epsilon^{3/2}}})$, where $d$ is a constant representing the dimension of the model parameters. Detailed proofs are provided in Sec. \ref{appendix:1}.
\end{theorem}

\section{Experiment}

Building upon the experimental settings of prior coreset selection work \cite{hao2023bilevel}, the proposed F$^2$CTO is evaluated on rehearsal-based continual learning tasks, with the evaluation further extended to distributed scenarios. We compare F$^2$CTO with a range of state-of-the-art methods, including the distributed trilevel optimization methods AFTO \cite{jiao2024provably} and DTZO \cite{jiaodtzo}, and the distributed coreset selection methods FedCS \cite{hao2025fedcs} and GCFL \cite{sivasubramanian2024gradient}. Additionally, we adapt several centralized coreset selection methods to the distributed setting and include them as additional baselines, including ACS \cite{dolatabadi2023adversarial}, BCSR \cite{hao2023bilevel}, and Greedy Coreset \cite{borsos2020coresets}. The experiments are conducted on a server equipped with two NVIDIA GeForce RTX 5090 GPUs. All models are implemented in PyTorch, and the detailed experimental settings are summarized in Table \ref{table:settings}. In our experiments, we aim to answer the following questions:
\textbf{(Q1)} Can the proposed F$^2$CTO achieve superior performance in distributed robust coreset selection? \textbf{(Q2)} Is the optimization at each level of the proposed framework effective?
\textbf{(Q3)} Can the proposed framework be applied to large-scale settings?
\textbf{(Q4)} Is the proposed first-order algorithm more efficient than traditional algorithms?  \textbf{(Q5)} Can periodic communication be effectively integrated into the proposed framework to further improve efficiency?

\begin{figure}[t]
\centering\includegraphics[width=1\linewidth]{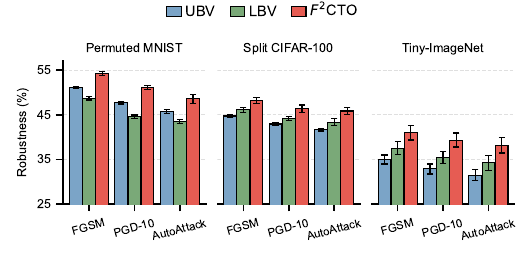}
\vspace{-8mm}
    \caption{The comparisons between F$^2$CTO and its two bilevel variants.}
    \label{fig:ablation}
\vspace{-3mm}
\end{figure}

\begin{figure}[t]
\centering
  \subfloat[Robustness against FGSM attack]{%
       \centering \includegraphics[width=0.40\textwidth]{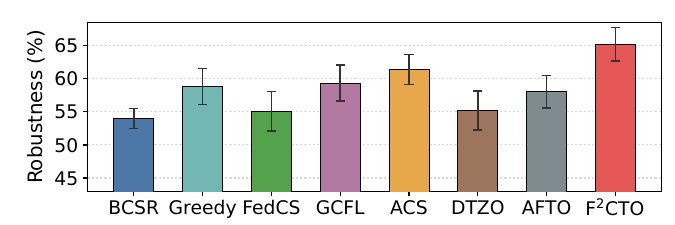}%
    }% 
\vspace{-3mm}\hspace{\textwidth} 
\centering   \subfloat[Robustness against PGD attack]{
        \includegraphics[width=0.40\textwidth]{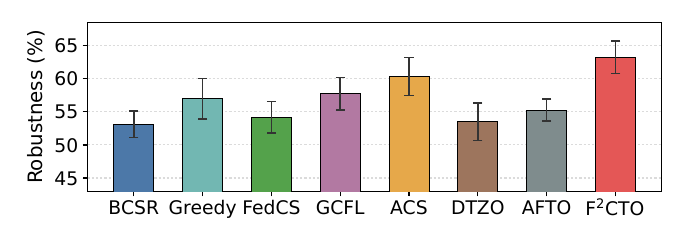}
    }    \caption{Results on large-scale IoT setting.}
    \label{fig:iot}
\end{figure}

\subsection{Main Results (Answering Q1)}

Following the experimental setup in \cite{hao2023bilevel}, we conduct experiments on Split CIFAR-100, Permuted MNIST, and Tiny-ImageNet. Two separate models are employed for robust coreset selection and model training, respectively. Consistent with \cite{hao2023bilevel}, we use the MLP for Permuted MNIST and ResNet-18 for the other datasets. Moreover, we use the average adversarial robustness across all tasks as an evaluation metric, and we assess adversarial robustness against three representative attacks: FGSM, PGD-10, and AutoAttack, following \cite{you2023beyond}. The comparisons between the proposed F$^2$CTO and the state-of-the-art methods across various datasets are presented in Table \ref{tab:experiment1}, it is seen that the F$^2$CTO outperforms all compared methods. This improvement can be attributed to two main factors: (1) In contrast to the state-of-the-art coreset selection methods (FedCS, GCFL, ACS, BCSR, and Greedy Coreset), this is the first work to unify robust optimization, coreset selection, and distributed learning. The resulting trilevel optimization framework effectively addresses robust coreset selection in a distributed manner, yielding improved performance. (2) Compared with the state-of-the-art distributed trilevel optimization methods AFTO and DTZO, which are designed for unconstrained trilevel optimization, the proposed F$^2$CTO is the first distributed method tailored to constrained trilevel optimization and can effectively handle level-wise constraints.

\subsection{Ablation Study (Answering Q2)}

This work introduces a trilevel optimization framework for distributed robust coreset selection. Within this framework, the upper two levels and the lower two levels each constitute a bilevel optimization problem. To assess the necessity of each optimization level, we conduct an ablation study with two bilevel variants. The upper-bilevel variant (UBV) retains the upper two levels while removing the third-level optimization, whereas the lower-bilevel variant (LBV) retains the lower two levels while removing the first-level optimization. Since LBV does not update data weights during optimization, a greedy-based strategy is adopted for coreset selection. It is worth noting that the second level cannot be removed independently, as doing so would disrupt the nested optimization structure. As shown in Fig. \ref{fig:ablation}, the proposed F$^2$CTO consistently outperforms both UBV and LBV across all datasets, demonstrating the effectiveness of the complete trilevel optimization framework.

\subsection{Scalability (Answering Q3)}

To evaluate the scalability of the proposed method in large-scale IoT scenarios, we conduct experiments on the Edge-IIoTset dataset \cite{ferrag2022edge} in a distributed network involving 200 workers. As shown in Fig. \ref{fig:iot}, the proposed F$^2$CTO maintains stable and superior performance compared with the baseline methods as the network size increases. These results demonstrate the scalability of F$^2$CTO and suggest that the proposed framework is applicable not only to cross-silo scenarios but also to large-scale cross-device environments.

\begin{figure}[t]
\centering\includegraphics[width=0.94\linewidth]{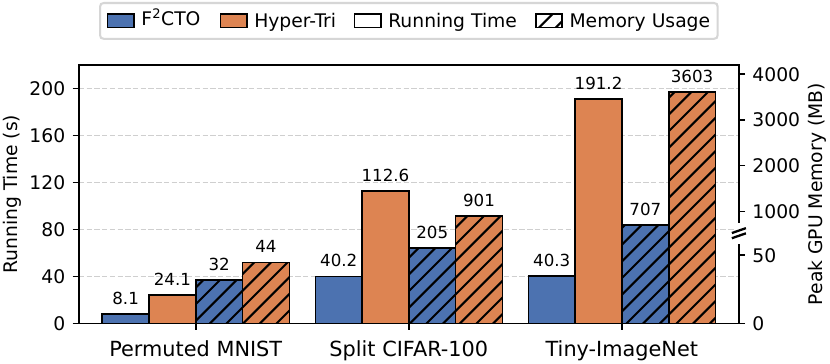}
\vspace{-2.5mm}
    \caption{Comparisons between the proposed F$^2$CTO and the hypergradient-based method in terms of running time (100 iterations) and memory usage.}
    \label{fig:running_time}
\vspace{-3mm}
\end{figure}

\subsection{Running Time and Memory Usage (Answering Q4)}

Compared with traditional hypergradient-based approaches to trilevel optimization \cite{giovannelli2025stochastic}, the proposed F$^2$CTO is a \textit{first-order} distributed algorithm that avoids computing hypergradients at each iteration, thereby reducing computational overhead. To empirically evaluate its computational efficiency, we compare the running time and memory usage of F$^2$CTO with the hypergradient-based trilevel method. As shown in Fig. \ref{fig:running_time}, F$^2$CTO consistently achieves shorter running time and lower memory usage, demonstrating its superior efficiency.

\begin{figure}[t]
     
  $\,\!$  \subfloat[Permuted MNIST\label{fig:sub1}]{%
        \includegraphics[width=0.48\textwidth]{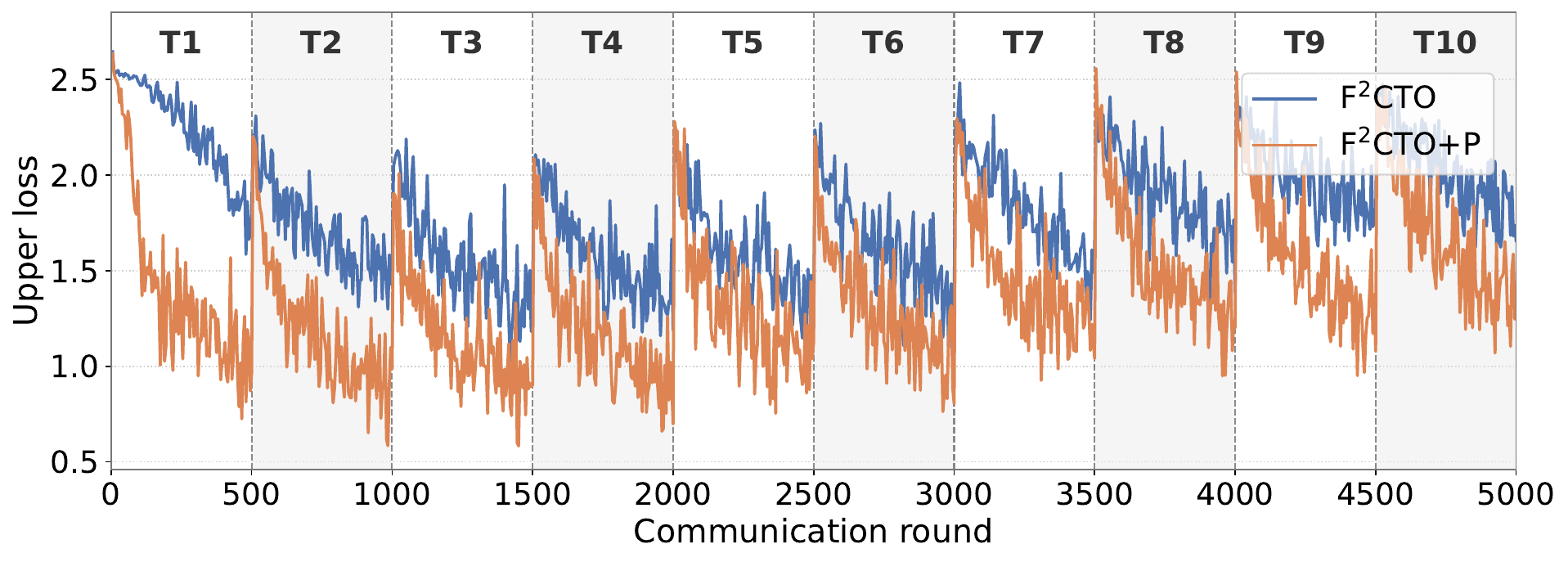}%
    }%
    \hfill \vspace{-1mm}
    \subfloat[Split CIFAR-100\label{fig:sub2}]{
        \includegraphics[width=0.48\textwidth]{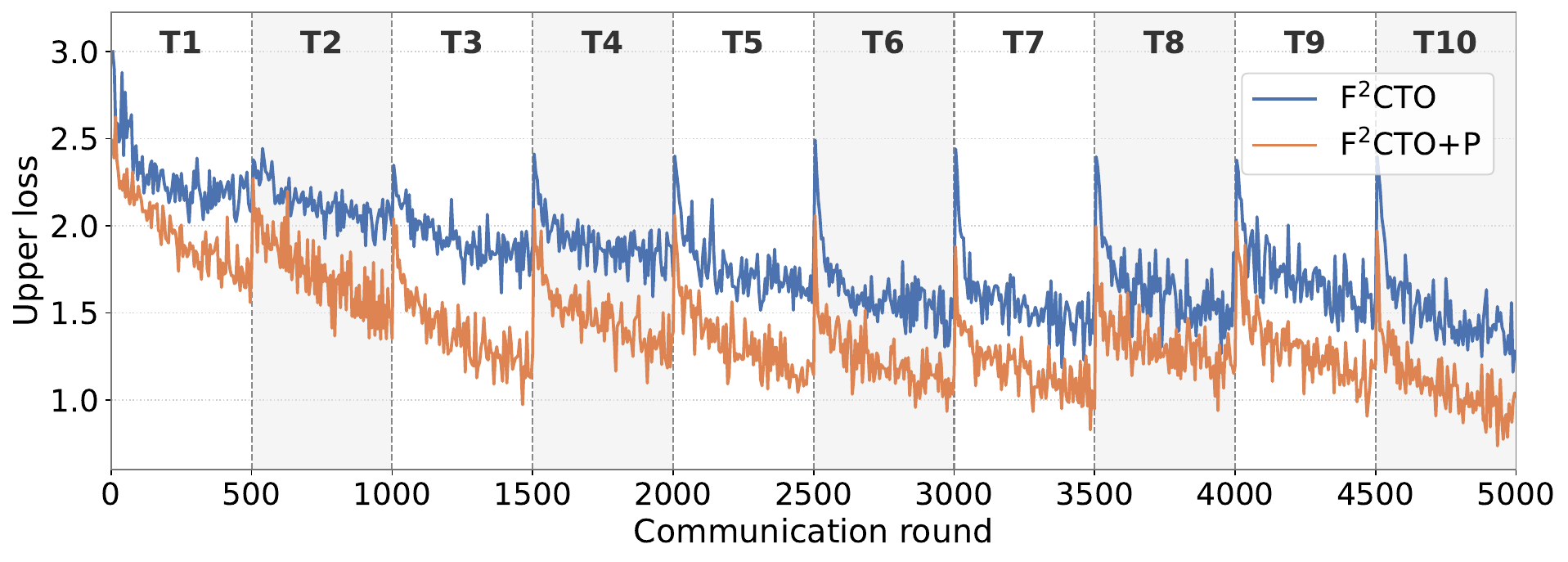}
    }    
    \hfill  \vspace{-2mm}
    \subfloat[Tiny-ImageNet\label{fig:sub3}]{
        \includegraphics[width=0.48\textwidth]{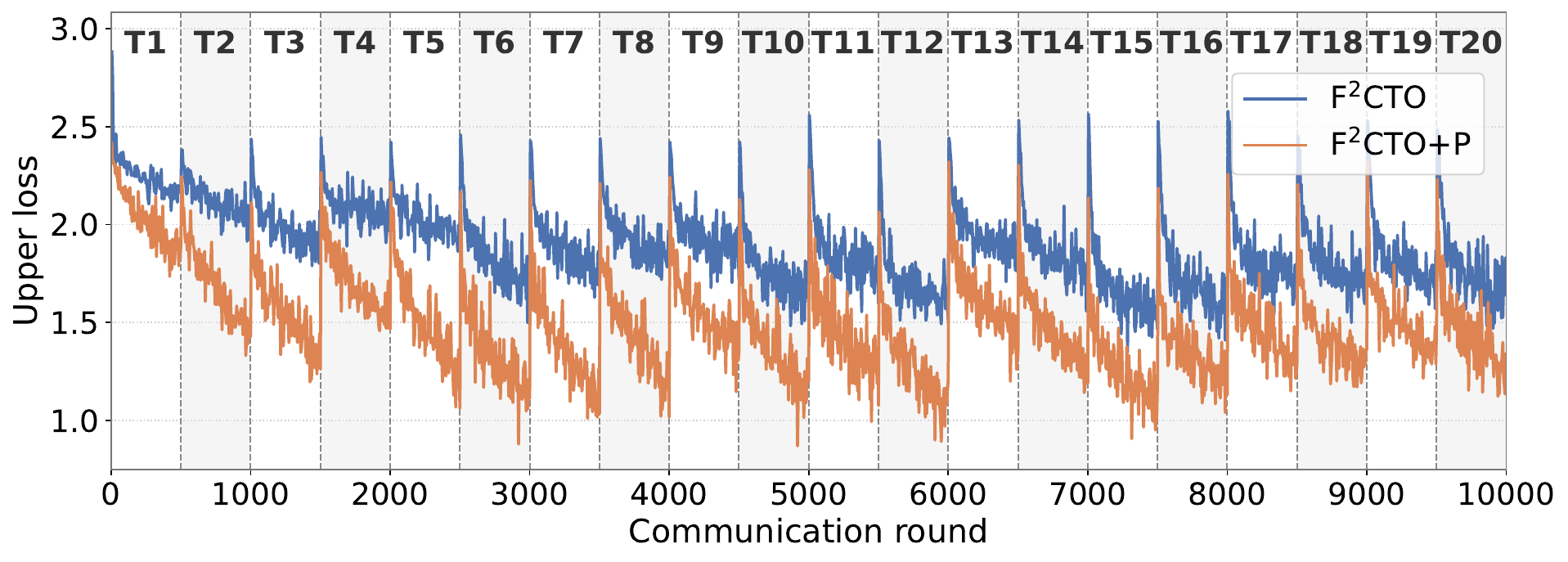}
    }
    \caption{F$^2$CTO with periodic communication strategy ($I=5$).}
    \label{fig:periodic}
\end{figure}

\subsection{Periodic Communication (Answering Q5)}

To further improve the communication efficiency of F$^2$CTO, we incorporate a periodic communication strategy, yielding F$^2$CTO+P. Instead of transmitting local model parameters to the master after every execution of the alternating projected gradient updates in Eqs. (\ref{eq:7_2_11})-(\ref{eq:7_6_14}), each worker performs $I$ rounds of local updates between two consecutive communication rounds, where $I\ge1$ is a pre-specified constant. As shown in Fig. \ref{fig:periodic}, F$^2$CTO+P achieves convergence with fewer communication rounds, indicating that periodic communication can be effectively integrated into the proposed framework.

\begin{table}[t]
\caption{Detailed experimental settings.}
\renewcommand{\arraystretch}{1.4}
\renewcommand{\tabcolsep}{1pt}
\centering
\scalebox{0.88}{
\begin{tabular}{l|ccccccccccccc}
\toprule
Dataset
& $N$
& $\eta_{\boldsymbol{\alpha}}$
& $\eta_{\boldsymbol{q}}$
& $\eta_{\boldsymbol{w}}$
& $\eta_{\boldsymbol{p}}$
& $c_1$
& $c_2$
& $c_3$
& $\varepsilon_{\rm{F}}$
& $\varepsilon_{\rm{P}}$
& $\varepsilon_{\rm{A}}$
& $R$
& $\hat{R}$ \\
\hline
S-CIFAR
& 10
& 5e-2
& 4e-3
& 5e-2
& 4e-3
& $4/255$
& 5
& $4/255$
& $4/255$
& $4/255$
& $4/255$
& 10
& 1 \\

P-MNIST
& 5
& 2e-2
& 8e-3
& 5e-3
& 8e-3
& $40/255$
& 5
& $40/255$
& $25/255$
& $40/255$
& $20/255$
& 10
& 1 \\

T-ImageNet
& 10
& 5e-2
& 4e-3
& 5e-2
& 4e-3
& $4/255$
& 5
& $4/255$
& $4/255$
& $4/255$
& $4/255$
& 10
& 1 \\

Edge-IIoT
& 200
& 5e-2
& 1e-2
& 5e-2
& 1e-2
& 5e-2
& 5
& 5e-2
& 1e-1
& 1e-1
& --
& 40
& 3 \\
\bottomrule
\end{tabular}
}

\label{table:settings}
\begin{flushleft}
\footnotesize
Here, $N$ is the number of workers in distributed systems,  $\eta_{\boldsymbol{\alpha}}$, $\eta_{\boldsymbol{q}}$, $\eta_{\boldsymbol{w}}$, $\eta_{\boldsymbol{p}}$ are step-sizes. $\varepsilon_{\rm{F}}$, $\varepsilon_{\rm{P}}$, and $\varepsilon_{\rm{A}}$ are perturbation budgets for FGSM, PGD-10, and AutoAttack. ${R}$ and $\hat{R}$ are the parameters for value-function refinement.
\end{flushleft}
\vspace{-4mm}
\end{table}

\section{Conclusion}
In this work, we explore the fundamental coupling among coreset selection, robust optimization, and distributed learning, formulating the distributed robust coreset selection as a trilevel optimization problem with level-wise constraints. To effectively address this, we propose F$^2$CTO, a federated first-order constrained trilevel optimization framework. Specifically, a hierarchical composite value-function reformulation is first introduced in F$^2$CTO to handle the trilevel structure, followed by a distributed first-order alternating projected gradient descent algorithm to solve the resulting problem. Theoretically, we provide the non-asymptotic convergence guarantee for the proposed F$^2$CTO, demonstrating its iteration and communication complexities to achieve the $\epsilon$-stationary point. Extensive experimental results on reliable continual learning have demonstrated the superior performance of F$^2$CTO.

\section{Appendix: Proofs of Theorem \ref{theorem:1} and \ref{theorem:2}}
\label{appendix:1}

In this section, we present the \textit{key steps} in the proofs of Theorems~\ref{theorem:1} and \ref{theorem:2}. To facilitate this, we first establish Lemmas \ref{lemma:1} and \ref{lemma:2}. By combining the definition of ${{\mathcal{P}_{\boldsymbol{W}}}}$, the update rule in Eq. (\ref{eq:7_2_15}), and the Triangle inequality, we can obtain:
\vspace{-1.5mm}
\begin{equation}
\label{eq:6_25_9}
\begin{array}{l}
||{{\mathcal{P}_{\boldsymbol{W}}}}({\boldsymbol{w}_i^{t+1}}) - {{\boldsymbol{w}^{t + 1}}}||\\
 = ||{{\mathcal{P}_{\boldsymbol{W}}}}({{\boldsymbol{w}^t}} - {\eta _{\boldsymbol{w}}}{\nabla _{\boldsymbol{w}}}\mathcal{L}_i(\boldsymbol{\alpha}_i^{t + 1},\boldsymbol{q}_i^{t + 1},{{\boldsymbol{w}^t}},\boldsymbol{p}_i^t)) \\ - {{\mathcal{P}_{\boldsymbol{W}}}}({{\boldsymbol{w}^t}} \!-\! \frac{1}{N}\!\sum\nolimits_{i = 1}^N \! {{\eta _{\boldsymbol{w}}}{\nabla _{\boldsymbol{w}}}\mathcal{L}_i(\boldsymbol{\alpha}_i^{t + 1},\boldsymbol{q}_i^{t + 1},{{\boldsymbol{w}^t}},\boldsymbol{p}_i^t)} )||\\
 \overset{(a)}{\le} {\eta _{\boldsymbol{w}}}||{\nabla _{\boldsymbol{w}}}\mathcal{L}_i(\boldsymbol{\alpha}_i^{t + 1},\boldsymbol{q}_i^{t + 1},{{\boldsymbol{w}^t}},\boldsymbol{p}_i^t) \\ \quad- \frac{1}{N}\sum\nolimits_{i = 1}^N {{\nabla _{\boldsymbol{w}}}\mathcal{L}_i(\boldsymbol{\alpha}_i^{t + 1},\boldsymbol{q}_i^{t + 1},{{\boldsymbol{w}^t}},\boldsymbol{p}_i^t)} ||\\
 \overset{(b)}{\le} {\eta _{\boldsymbol{w}}}(||{\nabla _{\boldsymbol{w}}}\mathcal{L}_i(\boldsymbol{\alpha}_i^{t + 1},\boldsymbol{q}_i^{t + 1},{{\boldsymbol{w}^t}},\boldsymbol{p}_i^t)|| \\  \quad + \frac{1}{N}\sum\nolimits_{i = 1}^N {||{\nabla _{\boldsymbol{w}}}\mathcal{L}_i(\boldsymbol{\alpha}_i^{t + 1},\boldsymbol{q}_i^{t + 1},{{\boldsymbol{w}^t}},\boldsymbol{p}_i^t)||} )\\
 \overset{(c)}{\le} 2{\eta _{\boldsymbol{w}}}{B_{\boldsymbol{w}}},
\end{array}
\vspace{-1mm}
\end{equation}
where inequality (a) follows from the non-expansiveness property of the projection operator $\mathcal{P}_{\boldsymbol{W}}$, inequality $(b)$ is obtained by applying the Triangle inequality, and inequality $(c)$ holds based on Assumption \ref{assum:1}. This completes the proof of Lemma \ref{lemma:1}. By the optimality condition of the projection operator in Eq. (\ref{eq:7_2_11}), it follows that
\vspace{-1.5mm}
\begin{equation}
\label{eq:6_23_11}
\begin{array}{l}
\langle {\nabla _{\boldsymbol{\alpha}} }\mathcal{L}_i({\boldsymbol{\alpha}_i ^t},{\boldsymbol{q}_i^t},{{\boldsymbol{w}^t}},{\boldsymbol{p}_i^t}),{\boldsymbol{\alpha}_i ^{t + 1}} - \boldsymbol{\alpha}_i \rangle \! \le \! \frac{\langle {\boldsymbol{\alpha}_i ^{t + 1}} - {\boldsymbol{\alpha}_i ^t},\boldsymbol{\alpha}_i  - {\boldsymbol{\alpha}_i ^{t + 1}}\rangle}{{{\eta _{\boldsymbol{\alpha}} }}} .
\end{array}
\end{equation}

By substituting $\boldsymbol{\alpha}_i=\boldsymbol{\alpha}_i ^t$ into Eq. (\ref{eq:6_23_11}), we can obtain:
\vspace{-1.5mm}
\begin{equation}
\label{eq:6_23_12}
\begin{array}{l}
\langle {\nabla _{\boldsymbol{\alpha}} }\mathcal{L}_i({\boldsymbol{\alpha}_i ^t},{\boldsymbol{q}_i^t},{{\boldsymbol{w}^t}},{\boldsymbol{p}_i^t}),{\boldsymbol{\alpha}_i ^{t + 1}} \!-\! {\boldsymbol{\alpha}_i ^t}\rangle  \!\le \! - \frac{||{\boldsymbol{\alpha}_i ^{t + 1}} \!- {\boldsymbol{\alpha}_i ^t}|{|^2}}{{{\eta _{\boldsymbol{\alpha}} }}}.
\end{array}
\end{equation}

\vspace{-1.5mm}

In addition, according to the $L$-smoothness in Assumption \ref{assum:1} and the fact that $\mathcal{L}(\{{\boldsymbol{\alpha}_i ^{t + 1}}\},\{\boldsymbol{q}_i^t\},{{\boldsymbol{w}^t}},\{\boldsymbol{p}_i^t\})=\frac{1}{N}\sum_{i=1}^N \mathcal{L}_i({\boldsymbol{\alpha}_i ^{t + 1}},{\boldsymbol{q}_i^t},{{\boldsymbol{w}^t}},{\boldsymbol{p}_i^t})$, it follows that
\vspace{-1.5mm}
\begin{equation}
\label{eq:6_23_14}
\begin{array}{l}
\mathcal{L}(\{{\boldsymbol{\alpha}_i ^{t + 1}}\},\!\{\boldsymbol{q}_i^t\},{{\boldsymbol{w}^t}},\!\{\boldsymbol{p}_i^t\})\! \le \! \mathcal{L}(\{{\boldsymbol{\alpha}_i ^{t}}\},\!\{\boldsymbol{q}_i^t\},{{\boldsymbol{w}^t}},\{\boldsymbol{p}_i^t\})\\ - \frac{1}{N}\sum_{i=1}^N ( {\frac{1}{{{\eta _{\boldsymbol{\alpha}} }}} - \frac{L}{2}} )||{\boldsymbol{\alpha}_i ^{t + 1}} - {\boldsymbol{\alpha}_i ^t}|{|^2}.
\end{array}
\end{equation}

\vspace{-1.5mm}

For the variable $\boldsymbol{q}_i$, by combining the optimality condition of the projection operator in Eq. (\ref{eq:7_6_12}) with $L$-smoothness in Assumption \ref{assum:1}, we can obtain:
\vspace{-1.5mm}
\begin{equation}
\label{eq:7_1_22}
\begin{array}{l}
\mathcal{L}(\{{\boldsymbol{\alpha}_i ^{t + 1}}\},\{\boldsymbol{q}_i^{t + 1}\},{{\boldsymbol{w}^t}},\{\boldsymbol{p}_i^t\})\! \le \! \mathcal{L}(\{{\boldsymbol{\alpha}_i^{t + 1}}\},\{\boldsymbol{q}_i^t\},{{\boldsymbol{w}^t}},\{\boldsymbol{p}_i^t\})\\ - \frac{1}{N}\sum_{i=1}^N ( {\frac{1}{{{\eta _{\boldsymbol{q}} }}} - \frac{L}{2}} )||{\boldsymbol{q}_i ^{t + 1}} - {\boldsymbol{q}_i ^t}|{|^2}.
\end{array}
\end{equation}

\vspace{-1.5mm}

For the variable $\boldsymbol{w}$, the projection happens after aggregating local parameters in Eq. (\ref{eq:7_2_15}). According to the optimality condition of the projection operator and by substituting $\boldsymbol{w}={\boldsymbol{w}^t}$:
\begin{equation}
\label{eq:6_23_16}
\begin{array}{l}
\!\left\langle {\frac{\sum\nolimits_{i = 1}^N \!{\nabla _{\boldsymbol{w}}}\mathcal{L}_i(\boldsymbol{\alpha}_i^{t+1},\boldsymbol{q}_i^{t + 1},{{\boldsymbol{w}^t}},{\boldsymbol{p}_i^t})}{N},{{\boldsymbol{w}^{t + 1}}} - {{\boldsymbol{w}^t}}} \right\rangle \! \le \!  - \frac{||{{\boldsymbol{w}^{t + 1}}} - {{\boldsymbol{w}^t}}|{|^2}}{{{\eta _{\boldsymbol{w}}}}}.
\end{array}
\end{equation}

Combining Eq. (\ref{eq:6_23_16}) with Eq. (\ref{eq:7_2_19}), we  have:
\vspace{-1.5mm}
\begin{equation}
\label{eq:6_25_5}
\begin{array}{l}
    \mathcal{L}(\{\boldsymbol{\alpha}_i^{t+1}\},\{\boldsymbol{q}_i^{t + 1}\},{{\boldsymbol{w}^{t + 1}}},\{{\boldsymbol{p}_i^t}\}) \!\le\! -( {\frac{1}{{{\eta _{\boldsymbol{w}} }}} \!-\! \frac{L}{2}} )||{{\boldsymbol{w}^{t+1}}} - {{\boldsymbol{w}^t}}|{|^2}  \\ +\mathcal{L}(\{\boldsymbol{\alpha}_i^{t + 1}\},\{\boldsymbol{q}_i^{t + 1}\},{{\boldsymbol{w}^t}},\{{\boldsymbol{p}_i^t}\}).
\end{array}
\end{equation}

\vspace{-1.5mm}

Likewise, for  variable $\boldsymbol{p}_i$, according to the optimality condition of the projection operator in Eq. (\ref{eq:7_6_14}), we can get:
\vspace{-1.5mm}
\begin{equation}
\begin{array}{l}
\left\langle { {{\nabla _{\boldsymbol{p}}}} \mathcal{L}_i(\boldsymbol{\alpha}_i^{t + 1},\boldsymbol{q}_i^{t + 1},{{\mathcal{P}_{\boldsymbol{W}}}}({\boldsymbol{w}_i^{t+1}}),{\boldsymbol{p}_i^t}),\boldsymbol{p}_i^{t + 1} - \boldsymbol{p}_i^t} \right\rangle \\ \le  - \frac{1}{{{\eta _{\boldsymbol{p}}}}}||\boldsymbol{p}_i^{t + 1} - \boldsymbol{p}_i^t|{|^2}. 
\end{array}
\end{equation}

\vspace{-1.5mm}

For simplicity, let ${\mathcal{L}_i^{t+1}}=\mathcal{L}_i(\boldsymbol{\alpha}_i^{t + 1},\boldsymbol{q}_i^{t + 1},{{\boldsymbol{w}^{t + 1}}},\boldsymbol{p}_i^t)$ and ${\hat{\mathcal{L}}_i^{t+1}}=\mathcal{L}_i(\boldsymbol{\alpha}_i^{t + 1},\boldsymbol{q}_i^{t + 1},{{\mathcal{P}_{\boldsymbol{W}}}}({{\boldsymbol{w}_i^{t+1}}}),\boldsymbol{p}_i^t)$, we can obtain:
\vspace{-1.5mm}
\begin{equation}
\label{eq:7_6_31}
\begin{array}{l}
\langle {\nabla _{\boldsymbol{p}}}{\mathcal{L}_i^{t+1}},\boldsymbol{p}_i^{t + 1} - \boldsymbol{p}_i^t\rangle \\
 =\! \langle {\nabla _{\boldsymbol{p}}}{\hat{\mathcal{L}}_i^{t+1}},\boldsymbol{p}_i^{t + 1} \!-\! \boldsymbol{p}_i^t\rangle 
 +\! \langle {\nabla _{\boldsymbol{p}}}{\mathcal{L}_i^{t+1}} \!-\! {\nabla _{\boldsymbol{p}}}{\hat{\mathcal{L}}_i^{t+1}},\boldsymbol{p}_i^{t + 1} \!-\! \boldsymbol{p}_i^t\rangle \\
 \!\le \! - \frac{1}{{{\eta _{\boldsymbol{p}}}}}||\boldsymbol{p}_i^{t + 1} \!-\! \boldsymbol{p}_i^t|{|^2}
\!  +\! \langle {\nabla _{\boldsymbol{p}}}{\mathcal{L}_i^{t+1}} \!-\! {\nabla _{\boldsymbol{p}}}{\hat{\mathcal{L}}_i^{t+1}},\boldsymbol{p}_i^{t + 1} \!-\! \boldsymbol{p}_i^t\rangle .
\end{array}
\end{equation}

By combining Eq. (\ref{eq:7_6_31}) with the $L$-smoothness, i.e., 
$\mathcal{L}_i(\boldsymbol{\alpha}_i^{t + 1},\boldsymbol{q}_i^{t + 1},{{\boldsymbol{w}^{t + 1}}},\boldsymbol{p}_i^{t + 1}) \le {\mathcal{L}_i^{t+1}}
 + \langle {\nabla _{\boldsymbol{p}} }{\mathcal{L}_i^{t+1}},\boldsymbol{p}_i^{t + 1} - \boldsymbol{p}_i^t\rangle  + \frac{L}{2}||\boldsymbol{p}_i^{t + 1} - \boldsymbol{p}_i^t|{|^2}$ and ${\mathcal{L}}(\{\boldsymbol{\alpha}_i^{t + 1}\},\{\boldsymbol{q}_i^{t + 1}\},{{\boldsymbol{w}^{t + 1}}},\{\boldsymbol{p}_i^{t + 1}\})=\frac{1}{N}\sum_{i=1}^N\mathcal{L}_i(\boldsymbol{\alpha}_i^{t + 1},\boldsymbol{q}_i^{t + 1},{{\boldsymbol{w}^{t + 1}}},\boldsymbol{p}_i^{t + 1})$, we can obtain:
 \vspace{-1.5mm}
\begin{equation}
\label{eq:7_1_27}
\begin{array}{l}
\!\mathcal{L}(\{ \boldsymbol{\alpha}_i^{t + 1}\},\{\boldsymbol{q}_i^{t + 1}\},{{\boldsymbol{w}^{t + 1}}},\{ \boldsymbol{p}_i^{t + 1}\} ) \!\le \!- (\frac{1}{{{\eta _{\boldsymbol{p}}}}} \!-\! \frac{L}{2})\!\sum\limits_{i = 1}^N \! \frac{{||\boldsymbol{p}_i^{t + 1} - \boldsymbol{p}_i^t|{|^2}}}{N}  \\ + \mathcal{L}(\{ \boldsymbol{\alpha}_i^{t + 1}\},\{\boldsymbol{q}_i^{t + 1}\},{{\boldsymbol{w}^{t + 1}}},\{ \boldsymbol{p}_i^t\} ) 
\\ + \underbrace {\begin{array}{l}
\frac{1}{N}\sum\nolimits_{i = 1}^N\! {\langle {\nabla _{\boldsymbol{p}}}{\mathcal{L}_i^{t+1}} - {\nabla _{\boldsymbol{p}}}{\hat{\mathcal{L}}_i^{t+1}},\boldsymbol{p}_i^{t + 1} - \boldsymbol{p}_i^t\rangle } 
\end{array}}_{{E_1}}.
\end{array}
\end{equation}

For the term $E_1$ in Eq. (\ref{eq:7_1_27}), we can derive:
\vspace{-1.5mm}
\begin{equation}
\label{eq:6_25_10}
\begin{array}{l}
{E_1} = \frac{1}{N}\sum\nolimits_{i = 1}^N {\langle {\nabla _{\boldsymbol{p}}}{\mathcal{L}_i^{t+1}} - {\nabla _{\boldsymbol{p}}}{\hat{\mathcal{L}}_i^{t+1}},\boldsymbol{p}_i^{t + 1} \!-\! \boldsymbol{p}_i^t\rangle } \\
 \overset{(a)}{\le} \! \frac{1}{{2\beta N}}\!\sum\limits_{i = 1}^N \! {||{\nabla _{\boldsymbol{p}}}{\mathcal{L}_i^{t+1}} \!-\! {\nabla _{\boldsymbol{p}}}{\hat{\mathcal{L}}_i^{t+1}}|{|^2}} 
 \!+\! \frac{\beta }{2N}\!\sum\limits_{i = 1}^N \!||\boldsymbol{p}_i^{t + 1} \!-\! \boldsymbol{p}_i^t|{|^2}\\
 \overset{(b)}{\le}\! \frac{{{L^2}}}{{2\beta N}}\!\sum\limits_{i = 1}^N \! {||{{\boldsymbol{w}^{t + 1}}} \!-\! {{\mathcal{P}_{\boldsymbol{W}}}}({\boldsymbol{w}_i^{t+1}})|{|^2}}  \!+\! \frac{\beta }{2N}\!\sum\limits_{i = 1}^N\!||\boldsymbol{p}_i^{t + 1} \!-\! \boldsymbol{p}_i^t|{|^2}\\
 \overset{(c)}{\le} \! \frac{{2{L^2}{\eta _{\boldsymbol{w}}^2}{B_{\boldsymbol{w}}^2}}}{{\frac{1}{{{\eta _{\boldsymbol{p}}}}} - \frac{L}{2}}} + \frac{1}{2N}(\frac{1}{{{\eta _{\boldsymbol{p}}}}} - \frac{L}{2})\sum\nolimits_{i = 1}^N||\boldsymbol{p}_i^{t + 1} - \boldsymbol{p}_i^t|{|^2},
\end{array}
\vspace{-1mm}
\end{equation}
where inequality (a) is based on the Young's inequality, inequality (b) applies the $L$-smoothness, and inequality (c) holds due to Eq. (\ref{eq:6_25_9}). Substituting the upper bound of $E_1$ from Eq. (\ref{eq:6_25_10}) into Eq. (\ref{eq:7_1_27}) yields:
\vspace{-1.5mm}
\begin{equation}
\label{eq:6_24_11}
\begin{array}{l}
\mathcal{L}(\{ \boldsymbol{\alpha}_i^{t + 1}\} ,\{\boldsymbol{q}_i^{t + 1}\},{{\boldsymbol{w}^{t + 1}}},\{ \boldsymbol{p}_i^{t + 1}\} ) \le \frac{{2{L^2}{\eta _{\boldsymbol{w}}^2}{B_{\boldsymbol{w}}^2}}} {{{{{1/\eta _{\boldsymbol{p}}}}} - L/2}} \\ - (\frac{1}{{{\eta _{\boldsymbol{p}}}}} \!-\! \frac{L}{2}) \! \sum\limits_{i = 1}^N \! \frac{{||\boldsymbol{p}_i^{t + 1} - \boldsymbol{p}_i^t|{|^2}}}{2N} \!+ \!\mathcal{L}(\{ \boldsymbol{\alpha}_i^{t + 1}\} ,\{\boldsymbol{q}_i^{t + 1}\},{{\boldsymbol{w}^{t + 1}}},\{ \boldsymbol{p}_i^t\} ).
\end{array}
\end{equation}

Consequently, combining Eqs. (\ref{eq:6_23_14}), (\ref{eq:7_1_22}), (\ref{eq:6_25_5}), and (\ref{eq:6_24_11}) completes the proof of Lemma \ref{lemma:2}. Summing both sides of Eq. (\ref{eq:lemma2}) from $t=0$ to $t=T-1$, and using Definition \ref{def:1}, we get:
\vspace{-1.5mm}
\begin{equation}
\label{eq:6_23_32}
\begin{array}{l}
\sum\limits_{t = 0}^{T - 1}( \frac{{{\eta _{\boldsymbol{\alpha}} }}}{2}( {1 - \frac{{L{\eta _{\boldsymbol{\alpha}} }}}{2}} )\sum\limits_{i = 1}^N {||g_{\boldsymbol{\alpha} ,i}^t|{|^2}}+\frac{{{\eta _{\boldsymbol{q}} }}}{2}( {1 - \frac{{L{\eta _{\boldsymbol{q}} }}}{2}} )\sum\limits_{i = 1}^N {||g_{\boldsymbol{q} ,i}^t|{|^2}}  \\ +\! \frac{{{\eta _{\boldsymbol{w}}}}}{2}( {1 \!-\! \frac{{L{\eta _{\boldsymbol{w}}}}}{2}} )||g_{\boldsymbol{w}}^t||{^2} \!+\! \frac{{{\eta _{\boldsymbol{p}} }}}{2}( {1 \!-\! \frac{{L{\eta _{\boldsymbol{p}} }}}{2}} )\sum\limits_{i = 1}^N {||g_{\boldsymbol{p} ,i}^t|{|^2}})-\frac{{{L^2}\eta _{\boldsymbol{w}}^2B_{\boldsymbol{w}}^2T}}{{2( {\frac{1}{{{\eta _{\boldsymbol{p}}}}} - \frac{L}{2}} )}} \\
 \le \mathcal{L}(\{\boldsymbol{\alpha}_i ^0\},\{\boldsymbol{q}_i^0\},{\boldsymbol{w}^0},\{\boldsymbol{p}_i^0\}) - \mathcal{L}(\{\boldsymbol{\alpha}_i ^T\},\{\boldsymbol{q}_i^T\},{{\boldsymbol{w}^T}},\{\boldsymbol{p}_i^T\}).
\end{array}
\end{equation}

By the choice of the step-sizes, i.e.,
${\eta _{\boldsymbol{\alpha}} }={\eta _{\boldsymbol{q}}} = {\eta _{\boldsymbol{w}}} = {\eta _{\boldsymbol{p}}} = \eta  = {T^{ - 1/3}}$, when $T\ge L^3$, we can get $\frac{\eta }{2}( {1 - \frac{{L\eta }}{2}} ) \ge \frac{\eta }{4}( {1 - \frac{{L\eta }}{2}} ) \ge \frac{\eta }{8},  \frac{1}{\eta } - \frac{L}{2} \ge \frac{1}{{2\eta }}$. Thus, we can obtain:
\vspace{-1.5mm}
\begin{equation}
\label{eq:7_1_33}
\begin{array}{l}
\frac{1}{T}\sum\nolimits_{t = 0}^{T - 1} {||{G^t}||{^2}} 
 \\ \! = \frac{1}{T}\! \sum\nolimits_{t = 0}^{T - 1} \left( \sum\nolimits_{i = 1}^N \! \left( \vert{}\vert{}g_{\boldsymbol{\alpha} ,i}^t\vert{}\vert{}^2 + \vert{}\vert{}g_{\boldsymbol{q} ,i}^t\vert{}\vert{}^2 \!+\! \vert{}\vert{}g_{\boldsymbol{p},i}^t\vert{}\vert{}^2 \right) + \vert{}\vert{}g_{\boldsymbol{w}}^t\vert{}\vert{}^2 \right) \\
 \!\le \! \frac{{\mathcal{L}(\{\boldsymbol{\alpha}_i ^0\},\{\boldsymbol{q}_i^0\},{\boldsymbol{w}^0}\!,\{\boldsymbol{p}_i^0\}) - \mathcal{L}(\{\boldsymbol{\alpha}_i ^T\},\{\boldsymbol{q}_i^T\},{{\boldsymbol{w}^T}}\!,\{\boldsymbol{p}_i^T\})}}{{\frac{\eta }{4}( {1 - \frac{{L\eta }}{2}} )}T} \!+\! \frac{{2{L^2} {\eta ^2}B_{\boldsymbol{w}}^2}}{{\frac{\eta }{4}( {1 \!-\! \frac{{L\eta }}{2}} )( {\frac{1}{\eta } \!-\! \frac{L}{2}} )}}\\
\! \le \! \frac{{8( {\mathcal{L}(\{\boldsymbol{\alpha}_i ^0\},\{\boldsymbol{q}_i^0\},{\boldsymbol{w}^0}\!,\{\boldsymbol{p}_i^0\}) - \mathcal{L}(\{\boldsymbol{\alpha}_i ^T\},\{\boldsymbol{q}_i^T\},{{\boldsymbol{w}^T}}\!,\{\boldsymbol{p}_i^T\})} )}+ 32{L^2}{\eta ^3} T{ B_{\boldsymbol{w}}^2} }{\eta T} \\
\! \le \! \frac{{8( {\mathcal{L}(\{\boldsymbol{\alpha}_i ^0\},\{\boldsymbol{q}_i^0\},{\boldsymbol{w}^0},\{\boldsymbol{p}_i^0\}) - \mathcal{L}(\{\boldsymbol{\alpha}_i ^*\},\{\boldsymbol{q}_i^*\},{{\boldsymbol{w}^*}},\{\boldsymbol{p}_i^*\})} ) + 32{L^2} { B_{\boldsymbol{w}}^2} }}{{{T^{2/3}}}}.
\end{array}
\end{equation}

Consequently, combining Eq. (\ref{eq:7_1_33}) with the definition of an $\epsilon$-stationary point in Definition \ref{def:2} \textit{yields the iteration complexity result stated in Theorem \ref{theorem:1}.}

The communication complexity of the proposed method comprises the costs of value-function refinement and variable updates. In $(t+1)^{\rm{th}}$ iteration, each worker $i$ first locally updates the value-functions $V_{3,i}(\boldsymbol{\alpha}_i, \boldsymbol{w})$ and $V_{2,i}^{(1)}(\boldsymbol{\alpha}_i^t, \boldsymbol{w}^t)$ completely on-device, incurring zero communication cost. Then, to refine $V_2^{(2)}({\boldsymbol{\alpha}_i^t})$, each worker $i$ transmits $\boldsymbol{w}i^{Q} \in \mathbb{R}^d$ to the master, which aggregates them and broadcasts the global average $\boldsymbol{w}^{Q} = \frac{1}{N}\sum{i=1}^N \boldsymbol{w}_i^{Q} \in \mathbb{R}^d$ back to the workers.

Following this, the local variables are updated alternately, and each worker uploads $\boldsymbol{w}i^{t+1} \in \mathbb{R}^d$ to the master. The master then broadcasts the projected aggregate as the new global parameter, $\boldsymbol{w}^{t+1} = \mathcal{P}{\boldsymbol{W}}( \frac{1}{N}\sum_{i=1}^N \boldsymbol{w}i^{t+1} ) \in \mathbb{R}^d$. Multiplying the per-iteration communication cost by the iteration complexity $T(\epsilon)$ in Theorem \ref{theorem:1}, the overall communication complexity satisfies $C{\mathrm{comm}}(\epsilon) \le 128d \cdot T(\epsilon) = \mathcal{O}(\frac{d}{{\epsilon^{3/2}}})$, \textit{completing the proof of Theorem \ref{theorem:2}.}

\bibliographystyle{IEEEtran}
\bibliography{cas-refs}

\end{document}